\documentclass[10pt,twocolumn]{article}
\usepackage{style}
\newcommand{\cF}{\mathcal{F}}

\newcommand{\vx}{\vec x}

\newcommand{\va}{\vec a}

\newcommand{\vm}{\vec m}

\newcommand{\ve}{{\varepsilon}}
\newcommand{\expect}{{\mathbb{E}}}
 
\title{Dynamical Regimes of Diffusion Models}

\author[1]{Giulio Biroli%
\thanks{\href{mailto:giulio.biroli@ens.fr}{\texttt{giulio.biroli@ens.fr}}}}
\author[1]{Tony Bonnaire%
\thanks{\href{mailto:tony.bonnaire@ens.fr}{\texttt{tony.bonnaire@ens.fr}}}}
\author[2]{Valentin de Bortoli%
\thanks{\href{mailto:vdebortoli@google.com}{\texttt{vdebortoli@google.com}}}}
\author[3]{Marc M\'ezard%
\thanks{\href{mailto:marc.mezard@unibocconi.it}{\texttt{marc.mezard@unibocconi.it}}}}
\affil[1]{Laboratoire de Physique de l'Ecole Normale Sup\'erieure, ENS, Universit\'e PSL, CNRS, Sorbonne Universit\'e, Universit\'e Paris Cit\'e, F-75005 Paris, France}
\affil[2]{Computer Science Department, ENS, CNRS, PSL University}
\affil[3]{Department of Computing Sciences, Bocconi University, Milano, Italy}
\date{\vspace{-8ex}}
\begin{document}

\pagenumbering{arabic}
\twocolumn[
  \begin{@twocolumnfalse}
    \maketitle
    \selectlanguage{american}
    
    \begin{abstract}
    Using statistical physics methods, we study generative diffusion models in the regime where the dimension of space and the number of data are large, and the score function has been trained optimally. Our analysis reveals three distinct dynamical regimes during the backward generative diffusion process.
    The generative dynamics, starting from pure noise, encounters first a 'speciation' transition where the gross structure of data is unraveled, through a mechanism similar to symmetry breaking in phase transitions. It is followed at later time by a 'collapse' transition where the trajectories of the dynamics become attracted to one of the memorized data points, through a mechanism which is similar to the condensation in a glass phase. For any dataset, the speciation time  can be found from a spectral analysis of the correlation matrix, and the  collapse time can be found from the estimation of an 'excess entropy' in the data. The dependence of the collapse time on the dimension and number of data provides a thorough characterization of the curse of dimensionality for diffusion models.  
    Analytical solutions for simple models like high-dimensional Gaussian mixtures substantiate these findings and provide a theoretical framework, while extensions to more complex scenarios and numerical validations with real datasets confirm the theoretical predictions.
    \end{abstract}
    
    \vspace{1cm}
  \end{@twocolumnfalse}
]
\saythanks

\section*{Introduction}
Machine learning has recently witnessed thrilling advancements, especially in the realm of generative models. At the forefront of this progress are diffusion models (DMs), which have emerged as powerful tools for modeling complex data distributions and generating new realistic samples. They have become the state of the art in generating images, videos, audio or 3d scenes
{\cite{Sohl_Dickstein2015,Song2019,Song_Sohl-Dickstein2021,guth2022wavelet,yang2022diffusion,saharia2022photorealistic,bar2024lumiere,poole2022dreamfusion}}.  
Although the practical success of generative diffusion models is widely recognized, their full theoretical understanding remains an open challenge. 
Rigorous results assessing their convergence on finite dimensional data have been obtained in \cite{de2021diffusion,lee2022convergence,de2022convergence,benton2023linear,conforti2023score,chen2022sampling}. However,
realistic data live in high dimensional spaces, where interpolation
between datapoints should face the curse of dimensionality \cite{donoho2000high}. A thorough understanding of how generative models escape this curse is still lacking. This requires approaches able to take into account that the number and the dimension of the data are simultaneously very large. 
In this work we face this challenge using statistical physics methods which have been developed to study probability distributions, disordered systems and stochastic processes in high-dimensions \cite{mezard2009information, charbonneau2023spin,bonnaire2023high}.


Diffusion models work in two stages. In the forward diffusion one starts from a data point (e.g., an image) and gradually adds noise to it, until the image has become pure noise. In the backward process one gradually denoises the image using a diffusion in an effective force field (the ``score'')  which is learnt leveraging techniques from score matching \cite{hyvarinen2005estimation,vincent2011connection} and deep neural networks. In this study, we focus on diffusion models which are efficient enough to learn the exact empirical score, i.e. the one obtained by noising the empirical distribution of data. 
In practical implementations, this should happen when one performs a long training of a strongly over-parameterized deep network to learn the score, in the situation when the number of data is not too large.

Within this framework we develop a theoretical approach able to characterize the dynamics of diffusion models in the simultaneous limit of large dimensions and large dataset. We show that the backward generative diffusion process consists of three subsequent dynamical regimes.  
 The first one is basically pure Brownian motion. In the second one  the backward trajectory finds one of the main classes of the data (for instance if the data consists of images of horses and cars, a given trajectory will specialize towards one of these two categories). In the third regime, the diffusion ‘collapses’ onto one of the examples of the dataset: a given trajectory commits to the attraction basin of one of the data points, and the backward evolution brings it back to that exact data point. Since  diffusion models are defined as the time reversal of a forward noising process, the generative process has to collapse on the training dataset under the exact empirical score assumption \cite{cattiaux2021time,haussmann1986time}. We show, by performing a thorough analysis of the curse of dimensionality for diffusion models, that this memorization can be avoided at finite times only if the size of the dataset is exponentially large in the dimension. An alternative, which is the one used in practice, is to rely on regularization and approximate learning of the score, departing from its exact form.
Understanding this crucial aspect of generative diffusion is a key open challenge \cite{kadkhodaie2023generalization,yoon2023diffusion,cui2023analysis} for which analyzing what happens when the exact empirical score is used represents a first step.

Separating these three dynamical regimes, we identify two characteristic cross-over times. The  {\it speciation} time $t_S$ is the transition from pure noise  to the commitment of a trajectory towards one category. The {\it collapse} time $t_C$ is the time where the backward trajectory falls into the attractor of one given data point. We provide mathematical tools to predict these times in terms of structure of
data. We will first study simple models such as high-dimensional Gaussian mixtures, where we obtain a full analytical solution and hence a very detailed understanding. Within this setting, we show that in the simultaneous limit of large number and dimension of the data, the speciation and collapse cross-overs become sharp thresholds. Interestingly, both of them are related to phase transitions studied in physics. We  then extend our results to more general settings and discuss the key role played by the dimensionality of data and the number of samples. Finally, we  perform numerical experiments, and confront the theory to real data such as CIFAR-10, ImageNet, and LSUN, showing that our main findings hold in realistic cases. We  conclude highlighting the consequences and the guidelines offered by our results, and discussing the next research steps, in particular how to go beyond the ``exact empirical score'' framework.

\section*{Main Results and Setting}

We focus on cases in which the data can be organized in distinct classes. For simplicity, we consider below two classes,  identified in the spectrum of the covariance matrix of the data: this spectrum is assumed to display a single large eigenvalue along the principal component which we will denote $\Lambda$. This is a simplifying assumption; our analysis can be extended to more than two classes, and subclasses within classes. The data consists of $n$ data points 
$\vec a \in \mathbb{R}^d$. We assume that there exists a ``true'' underlying distribution $P_0(\va)$ from which data are drawn, and we denote by $P_0^e(\va)=\sum_{\mu=1}^n\delta(\va -\va_\mu)/n$, the empirical distribution of the data. 
The components of $\va$ are normalized to be finite\footnote{More precisely, we assume that the moments $\int d\vec a P_0(\vec a) a_i^p$ remain of order one for $d\to\infty$ and finite $p$.} for large $d$. This implies in particular that the expectation of $|\vec x|^2$ grows linearly with $d$.

There exist many variants of diffusion models which are basically equivalent. We focus here on the diffusion process which consists in $d$ independent Ornstein-Uhlenbeck Langevin equations,
\begin{align}\label{eq-langevin}
d\vx(t)=-\vx(t)dt+d\vec B(t),
\end{align}
where $d\vec B(t)$ is square root of two times the standard Wiener process (a.k.a., Brownian motion) in $\mathbb{R}^d$.
The ``exact empirical score" is given by $\cF_i(\vx,t)=\partial \log P_t^e(\vx)/\partial x_i$ where $P_t^e(\vx)$ is the noisy empirical distribution at time $t$ due to the process in \eqref{eq-langevin}
\begin{align}
P_t^e(\vx)=\int d\va \; P_0^e(\va) \frac{1}{\sqrt{2\pi
  \Delta_t}^d}\exp\left(-\frac{1}{2} \frac{(\vx-\va
  e^{-t})^2}{\Delta_t}\right).
\label{Pt_gen}
\end{align}
This is the convolution of the empirical distribution of the data, $P_0^e(\va)$, with a Gaussian law of variance
$\Delta_t=(1-e^{-2t})$. At long times $P_t^e(\vx)$ is a Gaussian distribution with zero mean and covariance equal to the identity. In DMs the generation of new data is obtained by time-reversing this process using the backward dynamics
\begin{align}\label{eq:tr}
-dy_i(t)=y_idt+2 \cF_i(y,t)dt +d\xi_i(t),
\end{align}
where the noise $d\xi_i(t)$ has the same distribution as in the forward
process.

\begin{figure}[!t]
\centering
\includegraphics[width=1\linewidth]{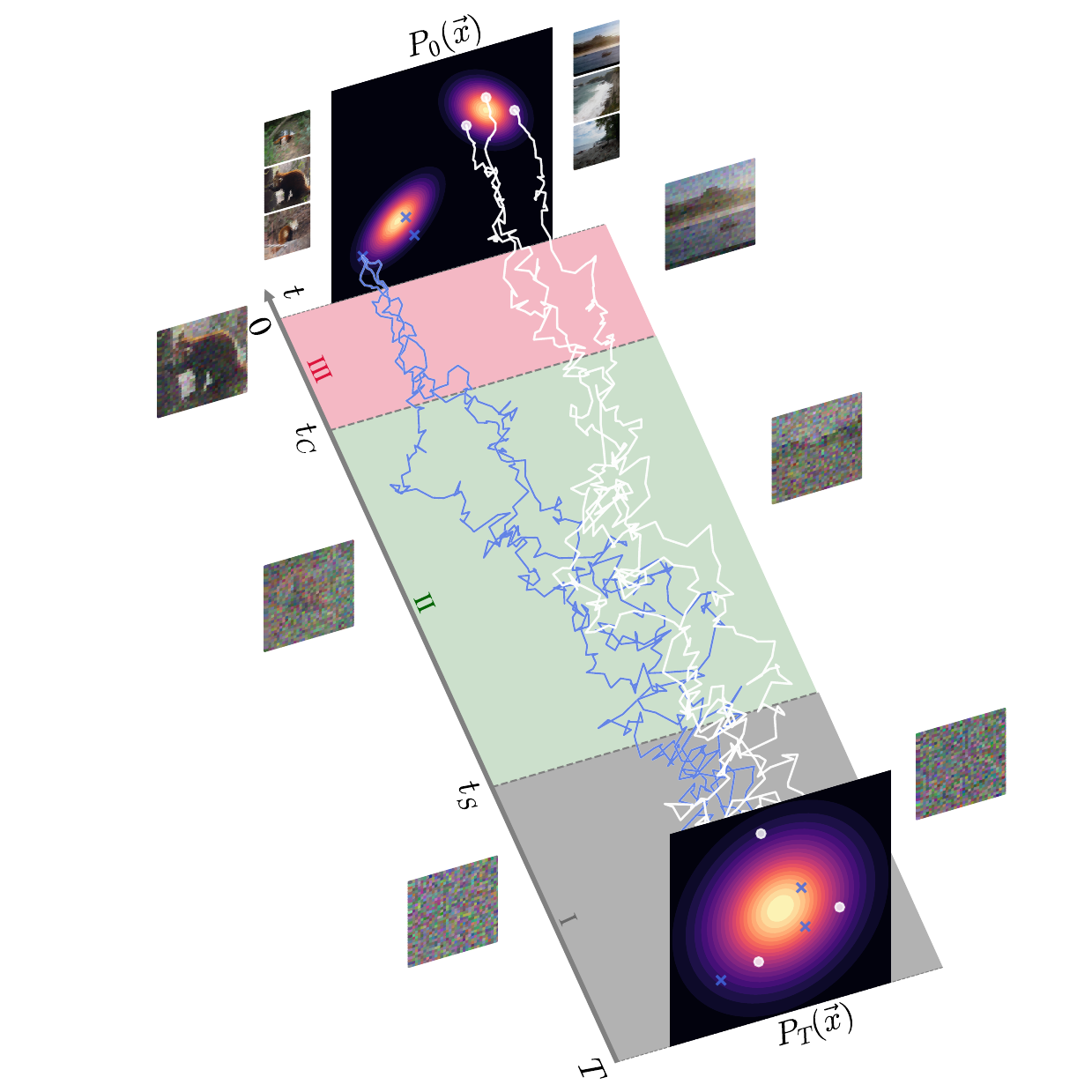}
\caption{Illustration of the three regimes of the backward dynamics though an example corresponding to a Gaussian mixture in two dimensions. Trajectories are colored white and blue according to their class at the end of the backward dynamics. In regime I, blue and white trajectories are fluctuating within the same bundle and $\vx$ is similar to white noise. At the speciation time $t_S$, the ensembles of blue and white trajectories divide and head towards the distribution associated to their respective class. Regime II is where the generative process constructs an $\vx$ which resembles to one element of the class (e.g., a seashore in the illustration) without being linked to any data of the training set. At the collapse time $t_C$, trajectories start to be attracted by the  data point on which they collapse at $t=0$. Regime III corresponds to memorization, whereas in regime I and II, the diffusion model truly generalizes. The images on the right and on the left are illustrations obtained from our ImageNet numerical experiment (notice the collapse on the panda and seashore from the training set at $t=0$).
}
\label{fig:recap}
\end{figure}

Our main contribution is the characterization of three dynamical regimes in the time-reversed process in the limit of large number of data and large dimension.
In {\it regime I}, at the beginning of the backward process, the random dynamical trajectories generated by \eqref{eq:tr} have not committed to a particular class of data. They have roughly the same probability to end up in one of the two classes. Figure~\ref{fig:recap} illustrates this behavior by showing that trajectories corresponding to different final classes are within a single bundle (or tube). In {\it regime II}, instead, the dynamical trajectories have committed to a particular class. In this case,  the trajectory will remain in the same class until the end of the backward process. 
During regime II, the backward dynamics generates the features needed to produce samples in a given class, but the fate of the trajectory in terms of class is sealed. In analogy with evolutionary dynamics, we call ``speciation'' the cross-over between regimes I and II, which has also important connections with the concept of symmetry breaking in physics \cite{GBM,ambrogioni}. As we shall show, the speciation cross-over takes place on a time-scale $t_S$ defined by
\begin{equation}\label{crit-spec}
    \Lambda e^{-2t_S} = 1,   \qquad \text{{\bf Speciation}: I}\rightarrow \text{II}
\end{equation}
where $\Lambda$ is the eigenvalue of the principal component of the covariance matrix of the data.  
Note that in the high-dimensional limit, if $\Lambda$ diverges with $d$ (typically one would expect $\Lambda \propto d$), then $t_S$ diverges logarithmically with $d$, and the speciation cross-over becomes a phase transition over time-scales of order $t_S$\footnote{Here, and in what follows, we measure time from the beginning of the forward process, i.e. large $t$ corresponds to the beginning of the backward process.}. 
We find that in the regimes I and II the DM generalizes (if the number of data is large enough), which means that the empirical distribution at time $t$, $P_t^e(\vx)$ is basically the same as the true $P_t(\vx)$, which is the convolution of $P_0$ and a Gaussian of variance $\Delta_t$. Therefore the distributions obtained by noising  the desired $P_0$ and the empirical $P_0^e$ are identical. \\
In {\it regime III}, the situation is completely different. There is no generalization; instead, the DM displays memorization of the training set. The probability distribution 
$P_t^e(\vx)$ no longer reflects $P_t(\vx)$. It actually decomposes in separated lumps around the points of the training set, and a given trajectory
is committed to the attractor of the original data point, that is reached at $t=0$.  Figure~\ref{fig:recap} illustrates this behavior.
We call ``collapse'' the cross-over between regimes II and III. 
As we shall show, the collapse takes place on a time-scale $t_C$ defined by
\begin{equation}\label{crit-coll}
  s(t_C)= s^{sep}(t_C),
\qquad \text{{\bf Collapse}: II}\rightarrow \text{III}
\end{equation}
where $s(t)=-\frac{1}{d} \int d\vx \; P_t(\vx) \log P_t(\vx)$ is the Shannon entropy per variable of the distribution at time $t$, and $s^{sep}(t) = \frac{\log n}{d} + \frac{1}{2} + \frac{1}{2}\log (2\pi \Delta_{t})$ is its counterpart for a mixture of $n$ Gaussian distributions  with variance $\Delta_t$ and centered on well separated points.
We shall argue below that the criterion \eqref{crit-coll} is actually valid beyond the ``exact empirical score'' hypothesis, and therefore provides a way to characterize the collapse (or its absence) in practical applications.  \\
For $t\gg 1$, the distribution $P_t(\vx)$ becomes a $d$-dimensional Gaussian, with entropy  $\frac{d}{2} + \frac{d}{2}\log (2\pi )$. In consequence, the difference between the two entropies,  
$f(t)=s^{sep}(t)-s(t)$, that we call ``excess entropy density'', equals $\frac{\log n}{d}$.
On the other hand, for $t\rightarrow 0$, the entropy $s(t)$ goes to the one of  $P_0(\va)$, while $\log (2\pi \Delta_{t})\to -\infty$. Therefore the excess entropy density goes to $-\infty$. When running the backward process, starting form large times where $f(t)=\frac{\log n}{d}>0$, this excess entropy density decreases and crosses $0$
at the collapse time $t_C$. Thus, by monitoring the time-dependence of $f(t)$, one can pinpoint when the collapse takes place. In numerical studies where $P_0$ is not known, one can approximate $f(t)$ as $f^e(t)=s^{sep}(t)-s^e(t)$, where $s^e(t)$ is the entropy per variable of the empirical distribution $P^e_t (\vx)$. This is a good approximation for $t\ge t_C$, where the two distributions $P^e_t$ and $P_t$ are identical and $f^e(t)=f(t)$. Instead, as we shall show, for $t<t_C$ the empirical excess entropy density vanishes while $f(t)<0$. 
The parameter $\alpha=\frac{\log n}{d}$ plays a key role in the collapse, as it allows to tune the value of $t_C$. A very small $\alpha$, implies a very small excess entropy density for $t\gg 1$. In this case the collapse takes place at the very beginning of the backward dynamics. One needs $\alpha \sim O(1)$ in order to have $t_C\sim O(1)$. To diminish $t_C$ (shrink regime III) and reduce the collapse, one has to increase $\alpha$. These findings characterize a curse of dimensionality which is different but related to the one arising in supervised learning: in order to avoid memorization, the number of data has to increase exponentially with the dimension $d$.
In practice, this unwanted  phenomenon is avoided by using an approximate score function which is smoother than the exact one, together with a large enough dataset. We will come back to this point later in the discussion. 

Our characterization of the backward dynamics is obtained using statistical physics methods developed to study phase transitions, disordered systems, and out-of-equilibrium dynamics in physics. We provide a brief introduction in the SI Appendix. The connection is not only methodological; in fact, our analysis unveils interesting relationships between the transition described above and phenomena intensively studied in physics in the last decades. In particular, the collapse transition is mutatis mutandis a glass transition in which the low energy glass states correspond to the training data.  The results we obtain in Eqs.~(\ref{crit-spec},\ref{crit-coll}), which are at the level of rigor of theoretical physics, provide guidelines and testable predictions for realistic applications. Numerical experiments on several subsets of realistic datasets (MNIST, CIFAR-10, ImageNet, and LSUN) confirm their validity.

\section*{Analysis of Gaussian mixtures}
An instructive simple example to study the backward dynamics in the large $d$ and $n$ limit is the Gaussian mixture model. In this case, the initial law $P_0(\vec a)$ is the superposition of two Gaussian clusters of equal weight, which we take for simplicity with means $\pm \vec m$, and the same variance $\sigma^2$.
We shall assume that  $|\vec m|^2 =d\tilde \mu^2$, with  $\sigma$ and $\tilde \mu$  of order 1. 

\begin{figure}
\centering
\includegraphics[width=1\linewidth]{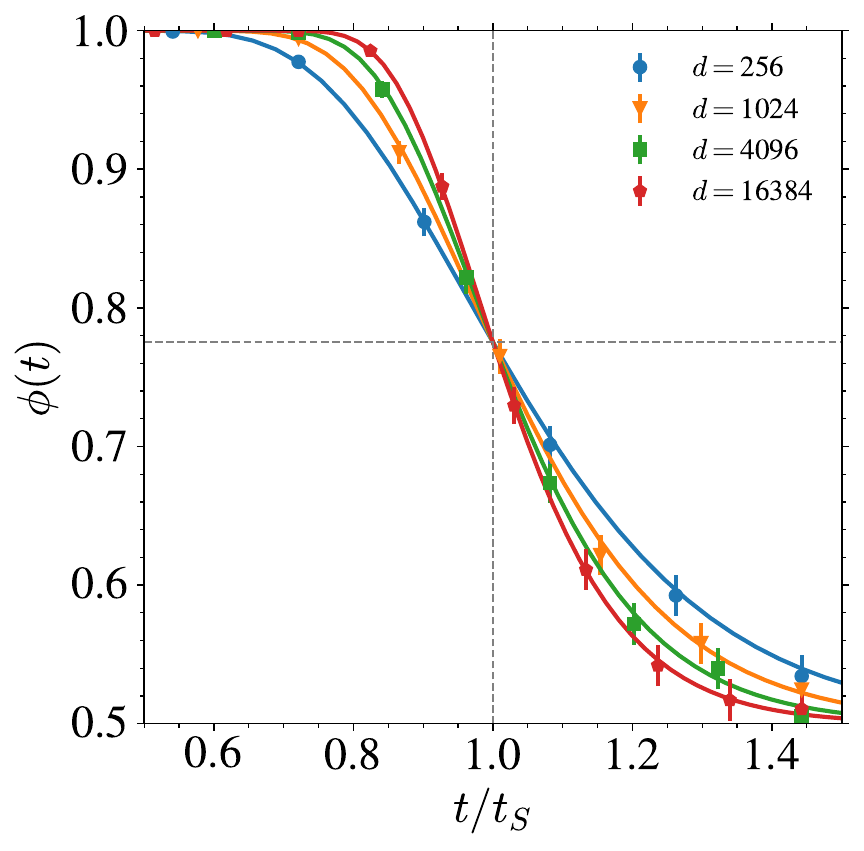}
\caption{Speciation in Gaussian mixtures: Evolution of $\phi(t)$ as a function of $t/t_S$ for several values of $d$ at fixed $\tilde{\mu} = 1$ and $\sigma = 1$. The solid line corresponds to the evaluation of \eqref{phi_GM} while the dots are obtained by sampling $10\,000$ clone trajectories. The vertical (resp. horizontal) dashed line corresponds to $t/t_S = 1$ (resp. $\phi(t) = 0.775$). Error bars correspond to thrice standard error.}
\label{fig:speciation_GM}
\end{figure}

{\bf Speciation time.} In order to show the existence of regimes I and II, and compute the speciation time, we focus on the following protocol which consists in cloning trajectories. We consider a backward trajectory starting at time $t_f\gg 1$ from a point $\vec x_f$ drawn from a random Gaussian distribution where all components are independent with mean zero and unit variance. This trajectory evolves backward in time, through the backward process until time $t<t_f$. At this time the trajectory has reached the point $\vec y(t)$, at which cloning takes place. 
One generates for $\tau<t$ two clones, starting from the same $\vec x_1(t)=\vec x_2(t) = \vec y(t)$, and evolving as independent trajectories $\vec x_1(\tau)$ and $\vec x_2(\tau)$, i.e. with independent thermal noises. We compute the probability $\phi(t)$ that the two trajectories ending in $\vec x_1(0)$ and $\vec x_2(0)$ are in the same class. Defining $P(\vec x_1, 0|\vec y,t)$ as the probability that the backward process ends in $\vec x_1$, given that it was in $\vec y$ at time $t$,  the joint probability of finding the trajectory in $\vec y$ at time $t$ and the two clones in $\vec x_1$ and $\vec x_2$ at time 0 is obtained as $Q(\vec x_1,\vec x_2,\vec y,t)=P(\vec x_1, 0|\vec y,t)P(\vec x_2, 0|\vec y,t)P(\vec y,t)$. Then $\phi(t)$ is the integral of $Q$ over $\vec x_1,\vec x_2,\vec y$ with the constraint $(\vec x_1 \cdot \vec m)(\vec x_2 \cdot \vec m) >0$. This simplifies into a one-dimensional integral (see SI Appendix):
\begin{align} \label{phi_GM}
\phi(t)=\frac{1}{2} \int_{-\infty}^{+\infty} dy \frac{G(y,m e^{-t},\Gamma_t)^2+G(y,-m e^{-t},\Gamma_t)^2 }{G(y,m e^{-t},\Gamma_t)+G(y,-m e^{-t},\Gamma_t)},
\end{align}
where $G(y,u,v)$ is a Gaussian probability density function for the real variable $y$, with mean  $u$ and variance $v$, and $m=|\vec m|=\tilde \mu \sqrt{d}$, $\Gamma_t= \Delta_t+\sigma^2 e^{-2t}$. The probability $\phi(t)$ that the two clones end up in the same cluster is a decreasing function of $t$, going from $\phi(0)=1$ to $\phi(\infty)=1/2$. In the large $d$ limit, the scaling variable controlling the change of $\phi$ is $\tilde \mu \sqrt{d}e^{-t}$ which can be rewritten as $\tilde \mu e^{-(t-t_S)}$ by using $t_S=(1/2)\log d$. This explicitly shows that speciation takes place at the timescale $t_S$ on a window of time of order one. As detailed in the SI Appendix, this expression for $t_S$ coincides with the one obtained from the general criterion \eqref{crit-coll}. We show in Fig.~\ref{fig:speciation_GM} the analytical result from \eqref{phi_GM} and direct numerical results obtained for increasingly larger dimensions. This comparison shows that our analysis is accurate already for moderately large dimensions. In the limit of infinite $d$, the analytical curve in Fig.~\ref{fig:speciation_GM} suddenly jumps from one to zero at $t/t_S=1$, corresponding to a symmetry breaking phase transition (or a threshold phenomenon) on the time-scale $t_S$. 
In the numerics, following finite size scaling theory \cite{privman1990finite}, we have defined the speciation time as the crossing point of the curves for different $d$, which corresponds approximatively to $\phi(t_S)=0.775$ and indeed is of the order $t_S= (1/2)\log d$ for $d\rightarrow \infty$. 
As it happens in mean-field theories of phase transitions \cite{opper2001advanced}, the large dimensional limit allows to obtain a useful limiting process. In our case, this leads to a full characterization of the asymptotic backward dynamics. At its beginning, i.e. in regime I, the overlap with the centers of the Gaussian model, $\pm \vec m\cdot\vec x(t)$, is of order $\sqrt{d}$. Defining $q(t)=\vec m\cdot \vec x(t) / \sqrt{d}$, one can obtain a closed stochastic Langevin equation on $q$ in an inverted potential $V(q, t)$ (see SI Appendix),
\begin{align}\label{eq:qt}
   -dq=-\frac{\partial V(q,t)}{\partial q}dt+d\eta(t),
\end{align}
where $\eta(t)$ is square root of two times a Brownian motion, and
\begin{align}
    V(q,t)=\frac{1}{2}q^2-2\tilde \mu^2 \log \cosh\left(q  e^{-t} \sqrt{d}\right).
\end{align}
At large $d$, this potential is quadratic at times $t\gg t_S=(1/2) \log d$, and it develops a double well structure, with a very large barrier, when $t\ll t_S=(1/2) \log d$. The trajectories of $q(t)$  are subjected to a force that drives them toward plus and minus infinity. The barrier between positive and negative values of $q$ becomes so large that trajectories commit to a definite sign of $q$: this is how the symmetry breaking takes place dynamically at the timescale $t_S$, in agreement with the previous cloning results.
Regime II corresponds to the scaling limit $q\rightarrow \infty$, where $\vec m \cdot \vec x(t)$ becomes of order $d$. In this regime the rescaled overlap $\vec m \cdot\vec x(t)/d$ concentrates, and its sign depends on the set of trajectories one is focusing on. Moreover, the stochastic dynamics of the $x_i$ corresponds to the backward dynamics for a single Gaussian centered in $\pm \vec m$. This shows that the dynamics generalizes, see SI Appendix (and also \cite{ghio2023sampling} for similar results).

\begin{figure}[!htb]
\centering
\includegraphics[width=1\linewidth]{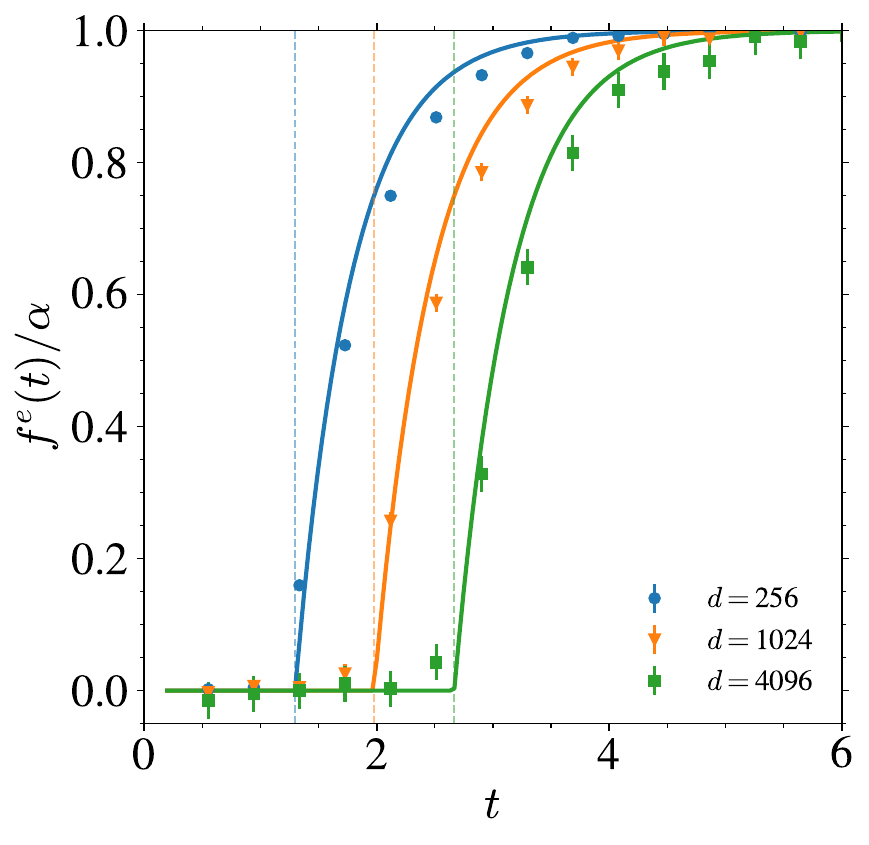}
\caption{Collapse in Gaussian mixtures: Evolution of the excess entropy density $f^e(t)/\alpha$ as a function of time $t$ for several values of $d$, at fixed $n=20\,000$. The solid lines are the theoretical predictions while the dots show the results of the numerical evaluation approximating the entropy from the dataset. The vertical dashed lines represent the collapse time $t_C$ predicted analytically for Gaussian mixtures given in \eqref{eq:tc_GM}. Error bars correspond to thrice the standard error.}
\label{fig:Collapse_GM}  
\end{figure}

{\bf Collapse time.}
In order to study the collapse, we consider the probability distribution $P_t^e(\vec x)$ given in \eqref{Pt_gen} around a point $\vec x$ which has been obtained in the forward process starting at $t=0$ from $\vec x=\vec a_1$. Our aim is to establish whether the score obtained from $P_t^e(\vec x)$ imposes a force that pushes trajectories toward $\vec a_1$ in the backward process, corresponding to memorization, or instead allows for generalization. We shall consider that both $n$ and $d$ go to infinity, keeping $\alpha=\frac{\log n}{d}$ fixed (it will be clear from the analysis that this is the correct ratio for the asymptotic analysis).\\  
The vector $\vec x$ we consider is equal to $\vec a_1 e^{-t}+\vec z \sqrt{\Delta_t}$ where $\vec z$ has iid Gaussian components with zero mean and unit variance. The probability can be written as 
$P_t^e(\vx)= \left[Z_1+Z_{2...n}\right]/\sqrt{2\pi\Delta_t}^d$ where $Z_1= e^{-\frac{1}{2}(\vx-\va_1 e^{-t})^2/(2\Delta_t)}= e^{-\frac{\vec z^2}{2}}$
and
\begin{align}
    Z_{2...n}= \sum_{\mu=2}^n e^{-\frac{1}{2}[(\vx-\va^\mu e^{-t})^2/(2\Delta_t)},
\label{Zrem}
\end{align}
In the large $d$ limit (to exponential accuracy), $Z_1\simeq e^{-d/2}$. The computation of $Z_{2...n}$ is instead tricky. Even though it is a sum of $n$ uncorrelated random contributions, standard concentration methods, e.g. central limit theorem, do not apply as each term of the sum corresponds to the exponential of a random variable scaling as $\log n$ \cite{ben2005limit}.  Statistical physics tools developed to study spin-glasses provide the method to solve this problem, see SI Appendix. In fact, given $\vx$, $Z_{2...n}$ is the partition function of a system with $n-1$ independent `random energy levels' $E^\mu =\frac{1}{2}\left[(\va_1-\va_\mu) e^{-t}+\vec z \sqrt{\Delta_t}\right]^2/(2\Delta_t)$ at thermal equilibrium at a temperature $1$ (the randomness comes from $\va_\mu$). This is some elaboration of the `Random Energy Model' which was introduced originally in \cite{derrida1981random} as a simple model of glass transition. A similar problem \cite{lucibello2023exponential} was studied recently in the related context of dense associative memories, using large-deviations and replica  methods (this connection was also noticed in \cite{ambrogioni_stat_thermo}). Using a similar approach we can show (see SI Appendix) that in the large $d$ and $n$ limit the distribution of $(1/d)\log Z_{2...n}$ concentrates on a value
$\psi_+ (t)$ which is an increasing function of the time $t$ and of $\alpha=\lim_{d,n\to\infty} (1/d)\log n$. The analytical computation shows the existence of a collapse time $t_C$ which separates two time regimes:
\begin{itemize}
     \item Regime III: at small  times, $t <t_C$, $\psi_+(t)<-(1/2)$ and the probability $P_t^e(\vx)$ is dominated by the term $Z_1= e^{-(\vx-\va_1 e^{-t})^2/(2\Delta_t)}$. When used in the backward diffusion, this gives a score that attracts $\vx$ towards $\va_1$ at short times. With probability one, the backward trajectory, starting at time $t$ from 
     $\vec x=\vec a_1 e^{-t}+\vec z \sqrt{\Delta_t}$, collapses at the end of the backward process on data point $\va_1$. In this regime the associated Random Energy Model is in a glass phase, which precisely corresponds to  memorization. 
 
     \item Regimes I and II: at large times, $t >t_C$, $\psi_+(t)>-(1/2)$ and the probability $P_t^e(\vx)$ is dominated by the term $Z_{2...n}$.
      This is the regime which is not collapsed and corresponds to generalization (regime II) or Brownian motion (regime I): in a typical point $\vec{x}(t)$, drawn from the population distribution $P_t(\vec{x})$, the empirical distribution $P_t^e(\vec{x},t)$ is equal to $P_t(\vec{x},t)$ at leading order in $d$. 
       In this regime the associated Random Energy Model is in the liquid (or high-temperature) phase. 
\end{itemize}
 The collapse time reads 
\begin{equation} \label{eq:tc_GM}
    t_{C}=\frac{1}{2}\log\left(1+\frac{\sigma^2}{n^{2/d}-1} \right).
\end{equation}
This equation makes manifest the curse of dimensionality: $t_C$ is of order one only when the number of data is exponential in the dimension. One needs $\frac{\log n}{d} \gg 1$ to push $t_C$ to zero and avoid the collapse. 
One can check that this equation for $t_C$ coincides with the one obtained from the general criterion \eqref{crit-coll}.
In Fig.~\ref{fig:Collapse_GM} we plot the empirical excess entropy density $f^e(t)$
from which one deduces $t_C$ as the largest time at which $f^e(t)=0$. 
The numerical results compare well with the analytical prediction, and confirms that the time-scale for collapse is well captured by our approach, even at moderately large values of $n$.
When $n,d \rightarrow \infty$ at fixed $\alpha = (1/d)\log n$, our analysis shows that regime III takes place on timescale of order one, and hence after the speciation transition, as illustrated in Fig.~\ref{fig:recap}.

\section*{Generalization to realistic datasets}
In the case of Gaussian mixtures we could thoroughly characterize the backward dynamics for large $n$ and $d$, using the knowledge of the initial distribution $P_0$. We now present a more general approach to the speciation and collapse transitions, for cases in which $P_0$ is not known or too complex to be analyzed exactly, but instead a dataset drawn from $P_0$ is available. In particular, we present arguments to establish criteria from Eqs.~(\ref{crit-spec},\ref{crit-coll}) which can be directly applied to realistic datasets.

The speciation transition can be analyzed using the covariance matrix, $ C_0$, of the initial data. Our assumption that data can be organized in two distinct and very different classes translates, in terms of $C_0$, in the existence of a strong principal component with eigenvalue $\Lambda$. The speciation transition can be generically understood in terms of the forward process: it corresponds to the time-scale at which the noise added to the data blurs the principal component, and hence the connection to a given class. On this timescale, the trajectories coming from different classes in the forward process coalesce within the same bundle, as illustrated in Fig.~\ref{fig:recap}. By time-reversal symmetry, it is therefore also on this time that the trajectories in the backward process separate and  commit to the different classes. 
By evaluating the covariance matrix of the noised data $x_i(t)$, one finds 
\begin{align}
    C(t)=C_0 e^{-2t}+\Delta_t {\mathbf I}.
\end{align}
The speciation time can be found comparing the two contributions on the RHS in the direction of the principal component of $ C_0$: $\Lambda e^{-2t}$ vs $\Delta_t$. The first one is associated to the fluctuations between different classes, whereas the second corresponds to the broadening of $P_t(\vec x)$ due to the noise. When the latter becomes of the same order of the former, the noise blurs the identification of trajectories in different classes. In consequence, one finds the criterion $\Lambda e^{-2t}\sim \Delta_t$, which shows that the variable controlling the speciation transition is indeed $\Lambda e^{-2t}$, as we have shown for the Gaussian mixture model.  Since for large $d$ one expects a large $\Lambda$, the asymptotic window over which the speciation transition takes place is when  
$\Lambda e^{-2t}$ is of order one. This leads to the general result stated in \eqref{crit-spec} \footnote{We have decided to associate the speciation transition with the time at which the variable $\Lambda e^{-2t}$ takes the value one, but this is a convention and the choice of another number of order one would work too when $\Lambda$ is large.}.
The argument above is obtained studying the time at which the structure of the data, present at $t=0$, is blurred by the noise. One can also tackle the problem starting by the other end, i.e. large times, and studying $P_t(\vec x)$ perturbatively in $e^{-t}$, which is a small parameter at the beginning of the backward process. By performing an expansion of (\ref{Pt_gen}) in $e^{-t}$, one finds that at leading order the distribution $P_t(\vec x)$ is a multivariate Gaussian with covariance $\frac{1}{\Delta_t}\left[{\mathbf I}-\frac{e^{-2t}}{\Delta_t}C_0\right]$. The expansion described above, and detailed in the SI, is similar to the Landau expansion of phase transitions \cite{LE}, in which symmetry breaking can be understood in terms of the instability of the quadratic part of $\log P_t(\vec x)$.
The speciation transition is like a symmetry breaking phase transition in $P_t(\vec x)$ \cite{GBM,ambrogioni}, and this instability happens for $\Lambda e^{-2t}=\Delta_t\simeq 1$. The general result \eqref{crit-spec} can thus be obtained from two different complementary perspectives. We expect it to hold in a broad set of cases. 

We now turn to the collapse transition. Generalization can be understood as the regime in which, at leading order in $d$, $P_t^e(\vec x)$ coincides with its population counterpart $P_t(\vec x)$ on typical noised data $\vec x(t)$ (i.e., drawn from $P_t(\vec x)$). This implies that at time $t$ the training set has been forgotten. At large $d$, and for a given data $\vec a$, the random vectors $\vec a e^{-t}+\vec z \sqrt{\Delta_t}$, drawn from $P_t^e(\vec x)$, lie with probability one on the ball\footnote{Actually they are located close to the circumference, as this is the region with the largest volume in large $d$.} of radius $\sqrt{d\Delta_t}$ around $\vec a e^{-t}$. At small time $t$, one can therefore envision the set ${\mathcal M}^e$ typically covered by the empirical distribution $P^e_t$ as the union of the non-overlapping balls centered around the data vectors $\va_\mu e^{-t}$. In this regime, the set ${\mathcal M}$ typically covered by the population distribution $P_t(\vec x)$ is clearly different. It is independent of the training set and not as singular for $t=0$. In consequence, in this regime $P_t^e(\vec x)$ is collapsed on the data and the diffusion model is in the memorization phase (regime III). 
By increasing $t$, the balls progressively grow and at a certain time, $t_C$, they cover the set ${\mathcal M}$. Beyond this point, one expects that the empirical and population distribution coincide on  ${\mathcal M}$ (at leading order in $d$). This picture suggests a volume argument to identify the collapse: one finds the time at which the volume of ${\mathcal M}^e$ coincides with the one of ${\mathcal M}$. The key point is that in large $d$, the volume $V_P$ covered by typical configurations associated to a given distribution $P$ scales as $V_P= e^{S}$ where $S=-\int d\vx P(\vx) \log P(\vx)$ is the Shannon entropy, which scales like $d$. The set ${\mathcal M}^e$ corresponds to $n$ d-dimensional Gaussian distributions with mean zero and covariance matrix  given by the identity times $\Delta_t$. For $t\le t_C$, the volume of ${\mathcal M}^e$ is the one of $n$ non-overlapping Gaussians, which therefore reads:
$  V_{{\mathcal M}^e}=n e^{S_G} $ where 
$  S_G=(d/2)(1+\log (2\pi \Delta_t)) $ is the entropy of one of the  Gaussian distributions. On the other hand, the volume of ${\mathcal M}$ is given by $e^{d s(t)}$. 
By requiring the equality of the two volumes, $V_{{\mathcal M}^e}=V_{{\mathcal M}}$, one finds the general result for $t_C$ given in \eqref{crit-coll}\footnote{Note that all the identities used in the volume argument are correct up to corrections exponentially small in $d$. For this reason, \eqref{crit-coll} is expressed in terms of intensive quantities, and hence valid up to vanishing corrections for $d\rightarrow \infty$.}. This criterion can also be obtained from the Random Energy Model method, generalizing our derivation for the Gaussian Mixture model. Note that the arguments above can also be applied to cases in which the score is learned by approximate models. In this case the distribution, $P_t(\vx)$ and its associated entropy density $s(t)$, have to be replaced by their model-dependent counterparts. \\
The arguments above offer another way, besides the exact computation by the Random Energy Model method, to illustrate the relationship between the collapse and the glass transition studied in physics \cite{berthier2011theoretical}. In fact, the ways in which $P^e_t(\vx)$ covers the space of $\vx$ before and after the collapse is the exact counterpart of how the Boltzmann law covers the configuration space before and after the ideal glass transition \cite{berthier2011theoretical}. In the glass phase the Boltzmann law is formed by lumps centered around amorphous optimal configurations (ideal glasses). Whereas in the liquid phase the Boltzmann law is spread over all configurations.
In the case studied here, the elements of the training set are the counterparts of the ideal glass configurations.

Having shown the generality of our criteria for speciation and collapse, we now test them on realistic data. 

\section*{Numerical experiments}
In the following we focus on various image datasets and we learn the score function using a state-of-the-art neural network with a finite number of samples $n$. We assume that, by using a heavily over-parameterized model, the resulting estimate $\hat{\mathcal{F}}$ of the score is close to the exact one.

{\bf Settings and datasets.} We train a diffusion model as first described in \cite{Ho2020}, and corresponding to \eqref{eq-langevin}, with a time horizon $T = 5.05$ and a linear schedule for the variance. The denoiser is a U-Net \cite{Ronneberger2015_unet} with an architecture similar to other traditional implementations of DMs \cite{Ho2020, song2020denoising, ambrogioni}. Our training sets are constructed by focusing on data divided in two classes and extracted from the image datasets: MNIST \cite{Lecun1998}, CIFAR-10 \cite{Krizhevsky2009_CIFAR}, downsampled ImageNet \cite{Chrabaszcz2017_Imagenet_downsampled}, and LSUN \cite{Yu2015_LSUN}. For example, in the case of ImageNet, the two classes we use are panda and seashore (as illustrated in Fig.~\ref{fig:recap}). The large variety of datasets allows us to explore different kinds of images and several values of $n$ and $d$, see Table~\ref{tab:datasets}. We refer to the SI Appendix for details on the processing of these datasets, the denoiser, and how the numerical experiments presented below are carried out.

\begin{table}[!t]
\centering
\caption{Datasets for speciation experiments}
\begin{tabular}{lcccccc}
Dataset name & $n$ & $d$ & $\Lambda$ & $t_S$ \\
\midrule
1. MNIST & 10\,000 & 1024 & 7.66 & 1.02 \\
2. CIFAR & 3000 & 3072 & 16.72 & 1.41 \\
3. ImageNet16 & 2000 & 768 & 3.05 & 0.56 \\
4. ImageNet32 & 2000 & 3072 & 12.11 & 1.25 \\
5. LSUN & 40\,000 & 12\,288 & 60.52 & 2.05 \\
\bottomrule
\label{tab:datasets}
\end{tabular}
\end{table}

\begin{figure}
\centering
\includegraphics[width=1\linewidth]{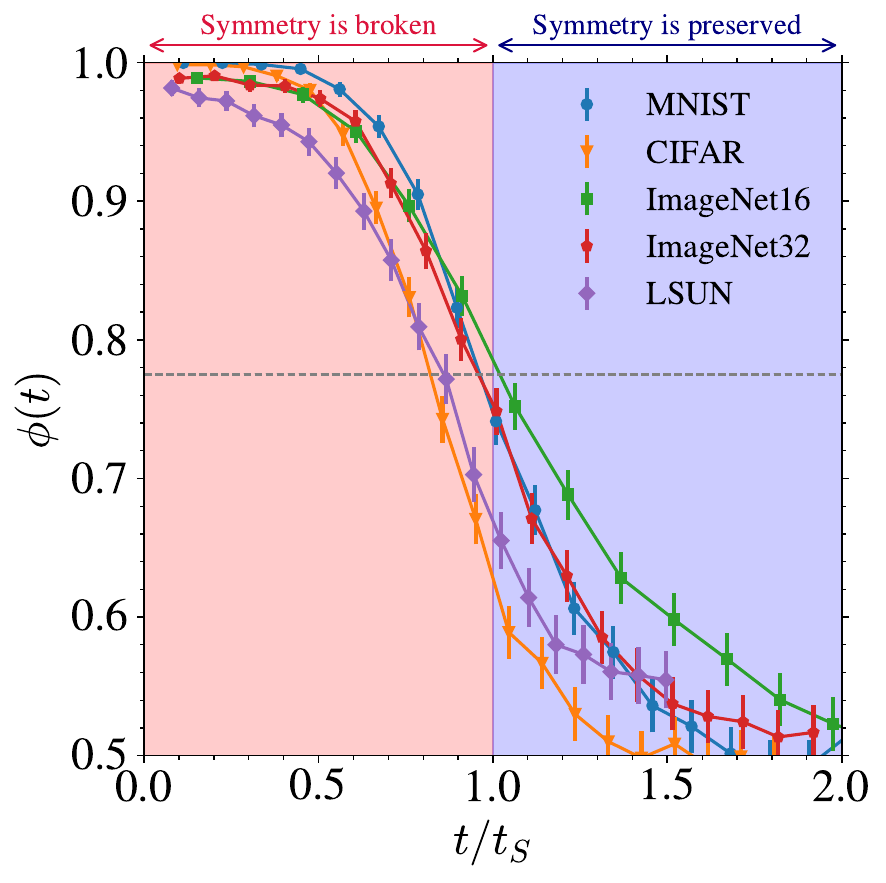}
\caption{Speciation in realistic datasets: Evolution of $\phi(t)$, the probability that the two clones end up in the same class, as a function of $t/t_S$ for several image datasets. The values of $t_S$ are the theoretical prediction for the speciation time obtained using \eqref{crit-spec} and listed in Table~\ref{tab:datasets}. The dashed horizontal line indicate $\phi(t) = 0.775$ and the errorbars correspond to thrice the standard error.
}
\label{fig:speciation}
\end{figure}

{\bf Speciation time.} To numerically extract the speciation time and compare it to the theoretical result given by~\eqref{crit-spec}, we use the same cloning procedure introduced for Gaussian mixtures. We generate two independent trajectories at time $t$ of the backward process, and numerically estimate the probability $\phi(t)$ that they end up in the same class by averaging over many initial conditions and backward processes. In order to recognize the class of $\vec x_1(0)$ and $\vec x_2(0)$, we use a classifier with a ResNet-18 architecture \cite{Kaiming2015_ResNet}, which has a test accuracy larger than $95\%$ on the datasets we focus on.
The corresponding results are shown in Fig.~\ref{fig:speciation}. The time axis has been rescaled by the theoretically predicted $t_S=(1/2)\log \Lambda$ from \eqref{crit-spec}. The values of $\Lambda$ and $t_S$ are listed in Table~\ref{tab:datasets}. Figure~\ref{fig:speciation} shows that indeed the speciation phenomenon is at play in realistic DMs: $\phi(t)$ goes from $0.5$ when $t \gg t_S$ to one when $t\ll t_S$, in a way which is qualitatively analogous to Fig.~\ref{fig:speciation_GM}. 
When rescaled with respect to $t_S$, the time-dependence exhibits remarkable similarity across vastly different image sets, suggesting evidence of the common underlying phenomenon of speciation.
Moreover, our prediction for the speciation time, $t_S$, captures well the behavior found in these numerical experiments. 

\begin{figure}[!t]
\centering
\includegraphics[width=1\linewidth]{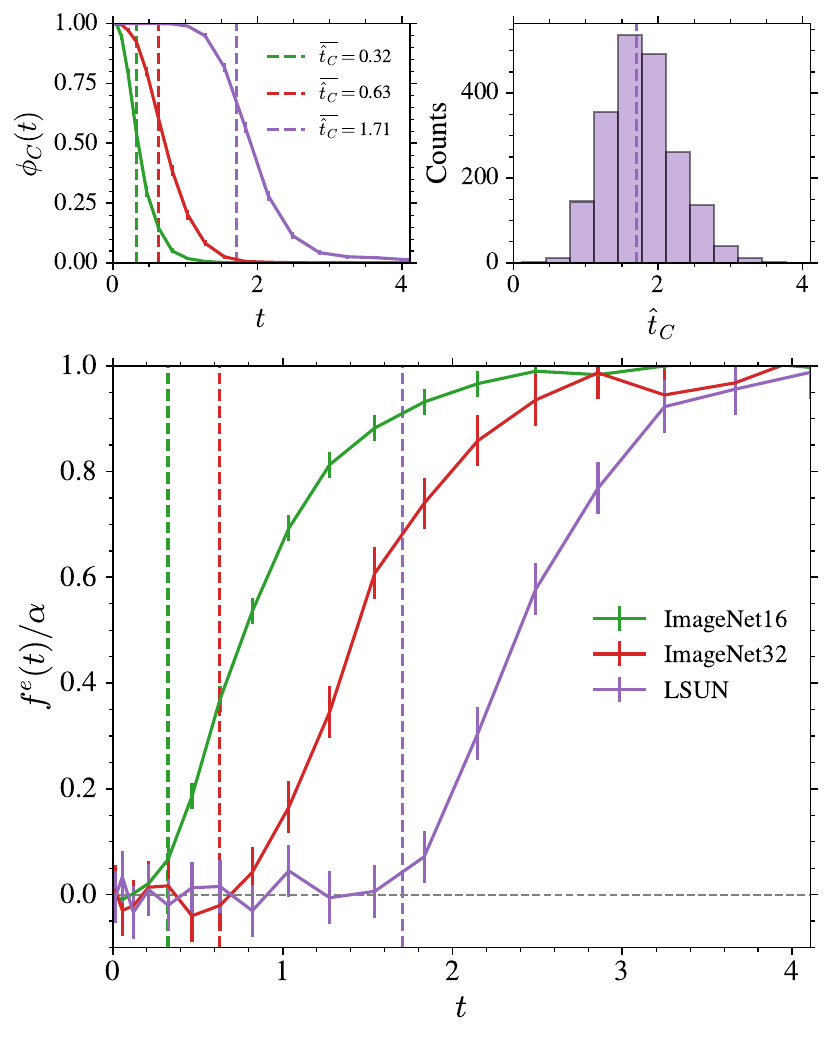}
\caption{Collapse in realistic datasets (ImageNet16, ImageNet32 and LSUN): \textit{(Top Left)} Evolution of $\phi_C(t)$, the probability that two cloned trajectories collapse on the same data of the training set at time zero. \textit{(Top Right)} Histograms of $\hat{t}_\mathrm{c}$ derived from the last-changing indices $\mu_\star$ on $4\,000$ generated samples for the LSUN dataset trained with $n=200$. \textit{(Bottom)} Evolution of the empirical excess entropy $f(t)/\alpha$. In all panels, the colored vertical dashed lines indicate the average of $\hat{t}_C$. The errorbars correspond to thrice the standard error.
}
\label{fig:collapse}
\end{figure}

{\bf Collapse time.} We focus on the datasets ImageNet16, ImageNet32, and LSUN. In order to be able to study thoroughly the collapse phenomenon, we have to keep the number of training data small ($n=200$ for LSUN and $n=2000$ for ImageNet). Otherwise, the model is not expressive enough to represent the singular behavior of the exact score at small times.
To estimate the collapse time numerically, and compare it to the theoretical result from \eqref{crit-coll}, we follow two distinct strategies. First, we use again the cloning procedure, but now we estimate the probability $\phi_C(t)$ that the two cloned trajectories collapse onto the same data of the training set at time zero.
The top left panel of Fig.~\ref{fig:collapse} shows the evolution of $\phi_C(t)$. The cross-over time where $\phi_C(t)$ goes from zero to one provides a first numerical estimate of $t_C$. An alternative estimation is found by tracking, during the backward process, the index of the closest neighbor in the dataset, noted $\mu_\star(t) \in \{1, \cdots, n\}$. For a single trajectory, we estimate the collapse time $\hat{t}_C$ as the last time during the backward process at which this index changes, meaning that for $t < \hat{t}_C$, the nearest neighbor of $\vec x(t)$ is always the same. The top right panel of Fig.~\ref{fig:collapse} represents the distribution of $\hat{t}_C$ measured in this way on the LSUN dataset for illustration. By computing the average of this distribution for each dataset, we obtain the estimates $\overline{\hat{t}_C}$ of the collapse time. We indicated them by vertical dashed lines in all panels of Fig.~\ref{fig:collapse}. The two different approaches to estimate the collapse time agree well, as seen in the top left panel where $\overline{\hat{t}_C}$ crosses $\phi_C(t)$ at around $0.60$ for all datasets.
We can now test the theoretical criterion for the collapse time given by~\eqref{crit-coll}, or equivalently by the largest time at which the empirical excess entropy density $f^e(t)$ is equal to zero, see the bottom panel of Fig.~\ref{fig:collapse}. The shape of $f^e(t)/\alpha$ is remarkably similar to the one obtained analytically for the Gaussian Model (Fig.~\ref{fig:Collapse_GM}). It clearly shows evidences of a transition.
The theoretical estimate, $f^e(t_C)=0$, also compares very well with the numerical ones $\overline{\hat{t}_C}$.

In conclusion, the numerical experiments presented in this section demonstrate the presence of speciation and collapse in realistic image datasets, and validate our theory, in particular the criteria from Eqs.~(\ref{crit-spec}) and~(\ref{crit-coll}) to identify the times at which these phenomena takes place.

\section*{Discussion}
In this work we have analyzed the backward dynamics of diffusion models assumed to be efficient enough to learn the exact empirical score. 
We have shown that for large number of data $n$, large dimension $d$, and in absence of regularization, the backward dynamics displays three different regimes. We have characterized the cross-overs between them, dubbed \emph{speciation} and \emph{collapse}, which become true transitions in the large $n,d$ limit. Interestingly, both of them have physical counterparts in the theory of phase transitions.
Speciation is a symmetry-breaking transition of $P_t(\vx)$ at which the most prominent classes are generated \cite{ambrogioni,GBM}. Collapse corresponds to a glass transition at which $P_t(\vx)$ fragments in an ensemble of lumps centered around the training data. 
Our approach characterizes the time at which speciation and collapse take place in terms of structure of data. The speciation time is determined by the eigenvalue of the principal component of the covariance matrix of the data, whereas the collapse time is governed by the entropy of the noised data distribution. \\
Although we have focused on a specific diffusion model, our results hold for the large variety of DMs which are based on inverting the noising process \cite{albergo2023stochastic}. These methods use different procedures to implement the denoising process $P_\infty(\vx) \rightarrow P_t(\vx)\rightarrow P_0 (\vx)$. 
However, as we have shown, the phenomena of speciation and collapse, can be understood and analyzed focusing on the properties of the distribution $P_t(\vx)$ alone.
In consequence, since all these different DMs lead to the same backward evolution of $P_t(\vx)$, the criteria from Eqs.~(\ref{crit-spec}) and~(\ref{crit-coll}) for speciation and collapse hold in general, whether the generative process is deterministic and flow-based or stochastic and diffusion-based.

Our work was done within the ``exact empirical score'' hypothesis. However, the phenomena we analyzed are relevant also when the score is learned approximately, as shown by our numerical experiments. Moving beyond the ``exact empirical score'' hypothesis opens up multiple avenues for further research. On the one hand, it would be interesting to analyze simple models of data and score in the limit of large number of data, dimension and parameters, as initiated in \cite{GBM,cui2023analysis}. On the other hand, it would be important to develop a quantitative study of the role of regularization on the phenomena we presented in this work. As discussed in \cite{GBM}, using a good model of the score is a key ingredient to generate data in the right proportions, i.e. reproducing the correct weights of the classes. In consequence, regularization methods could be detrimental in regime I and for speciation: by preventing the score to be close to the exact one they could lead to data with correct features (e.g., realistic images) but wrong proportions with respect to the training set. 
On the contrary, regularization is beneficial to avoid the collapse. As we have shown, diffusion models are ``cursed'': one needs an exponential number of data in $d$ to avoid the collapse on datapoints from the training set. Regularization allows to circumvent this problem in practice if the number of data is large enough \cite{kadkhodaie2023generalization} (see also SI Appendix). The volume argument we develop in this work can be applied beyond the ``exact empirical score'' hypothesis, and could offer a way to analyze quantitatively how the collapse depends on $n,d$ and the capacity of the model used to learn the score. 

Finally, our results also provide suggestions to improve and understand procedures used in practical applications. In particular, taking into account in practical implementations the existence of three regimes of the backward dynamics could lead to better performances.

\subsection*{Acknowledgments}
The authors thank F. Bach, S. Mallat, A. Montanari for helpful discussions. GB acknowledges support from the French government under the management of the Agence Nationale de la Recherche as part of the “Investissements d’avenir” program, reference ANR-19-P3IA0001 (PRAIRIE 3IA Institute). MM acknowledges the support of the PNRR-PE-AI FAIR project funded by the NextGeneration EU program.

\bibliography{pnas-sample}

\begin{thebibliography}{51}
\providecommand{\natexlab}[1]{#1}
\providecommand{\url}[1]{\texttt{#1}}
\expandafter\ifx\csname urlstyle\endcsname\relax
  \providecommand{\doi}[1]{doi: #1}\else
  \providecommand{\doi}{doi: \begingroup \urlstyle{rm}\Url}\fi

\bibitem[Albergo et~al.(2023)Albergo, Boffi, and Vanden-Eijnden]{albergo2023stochastic}
Michael~S Albergo, Nicholas~M Boffi, and Eric Vanden-Eijnden.
\newblock Stochastic interpolants: A unifying framework for flows and diffusions.
\newblock \emph{arXiv preprint arXiv:2303.08797}, 2023.

\bibitem[Ambrogioni(2023)]{ambrogioni_stat_thermo}
Luca Ambrogioni.
\newblock The statistical thermodynamics of generative diffusion models.
\newblock \emph{arXiv preprint arXiv:2310.17467}, 2023.

\bibitem[Bar-Tal et~al.(2024)Bar-Tal, Chefer, Tov, Herrmann, Paiss, Zada, Ephrat, Hur, Li, Michaeli, et~al.]{bar2024lumiere}
Omer Bar-Tal, Hila Chefer, Omer Tov, Charles Herrmann, Roni Paiss, Shiran Zada, Ariel Ephrat, Junhwa Hur, Yuanzhen Li, Tomer Michaeli, et~al.
\newblock Lumiere: A space-time diffusion model for video generation.
\newblock \emph{arXiv preprint arXiv:2401.12945}, 2024.

\bibitem[Ben~Arous et~al.(2005)Ben~Arous, Bogachev, and Molchanov]{ben2005limit}
G{\'e}rard Ben~Arous, Leonid~V Bogachev, and Stanislav~A Molchanov.
\newblock Limit theorems for sums of random exponentials.
\newblock \emph{Probability theory and related fields}, 132:\penalty0 579--612, 2005.

\bibitem[Benton et~al.(2023)Benton, De~Bortoli, Doucet, and Deligiannidis]{benton2023linear}
Joe Benton, Valentin De~Bortoli, Arnaud Doucet, and George Deligiannidis.
\newblock Linear convergence bounds for diffusion models via stochastic localization.
\newblock \emph{arXiv preprint arXiv:2308.03686}, 2023.

\bibitem[Berthier and Biroli(2011)]{berthier2011theoretical}
Ludovic Berthier and Giulio Biroli.
\newblock Theoretical perspective on the glass transition and amorphous materials.
\newblock \emph{Reviews of modern physics}, 83\penalty0 (2):\penalty0 587, 2011.

\bibitem[Biroli and M{\'e}zard(2023)]{GBM}
Giulio Biroli and Marc M{\'e}zard.
\newblock Generative diffusion in very large dimensions.
\newblock \emph{J. Stat. Mech.}, page 093402, 2023.

\bibitem[Bonnaire et~al.(2023)Bonnaire, Ghio, Krishnamurthy, Mignacco, Yamamura, and Biroli]{bonnaire2023high}
Tony Bonnaire, Davide Ghio, Kamesh Krishnamurthy, Francesca Mignacco, Atsushi Yamamura, and Giulio Biroli.
\newblock High-dimensional non-convex landscapes and gradient descent dynamics.
\newblock \emph{arXiv preprint arXiv:2308.03754}, 2023.

\bibitem[Cattiaux et~al.(2021)Cattiaux, Conforti, Gentil, and L{\'e}onard]{cattiaux2021time}
Patrick Cattiaux, Giovanni Conforti, Ivan Gentil, and Christian L{\'e}onard.
\newblock Time reversal of diffusion processes under a finite entropy condition.
\newblock \emph{arXiv preprint arXiv:2104.07708}, 2021.

\bibitem[Chaikin et~al.(1995{\natexlab{a}})Chaikin, Lubensky, and Witten]{LE}
Paul~M Chaikin, Tom~C Lubensky, and Thomas~A Witten.
\newblock \emph{Principles of condensed matter physics}, volume~10.
\newblock Cambridge university press Cambridge, 1995{\natexlab{a}}.

\bibitem[Chaikin et~al.(1995{\natexlab{b}})Chaikin, Lubensky, and Witten]{chaikin1995principles}
Paul~M Chaikin, Tom~C Lubensky, and Thomas~A Witten.
\newblock \emph{Principles of condensed matter physics}, volume~10.
\newblock Cambridge university press Cambridge, 1995{\natexlab{b}}.

\bibitem[Charbonneau et~al.(2023)Charbonneau, Marinari, Parisi, Ricci-tersenghi, Sicuro, Zamponi, and Mezard]{charbonneau2023spin}
Patrick Charbonneau, Enzo Marinari, Giorgio Parisi, Federico Ricci-tersenghi, Gabriele Sicuro, Francesco Zamponi, and Marc Mezard.
\newblock \emph{Spin Glass Theory and Far Beyond: Replica Symmetry Breaking after 40 Years}.
\newblock World Scientific, 2023.

\bibitem[Chen et~al.(2022)Chen, Chewi, Li, Li, Salim, and Zhang]{chen2022sampling}
Sitan Chen, Sinho Chewi, Jerry Li, Yuanzhi Li, Adil Salim, and Anru~R Zhang.
\newblock Sampling is as easy as learning the score: theory for diffusion models with minimal data assumptions.
\newblock \emph{arXiv preprint arXiv:2209.11215}, 2022.

\bibitem[Chrabaszcz et~al.(2017)Chrabaszcz, Loshchilov, and Hutter]{Chrabaszcz2017_Imagenet_downsampled}
Patryk Chrabaszcz, Ilya Loshchilov, and Frank Hutter.
\newblock A downsampled variant of imagenet as an alternative to the cifar datasets, 2017.

\bibitem[Conforti et~al.(2023)Conforti, Durmus, and Silveri]{conforti2023score}
Giovanni Conforti, Alain Durmus, and Marta~Gentiloni Silveri.
\newblock Score diffusion models without early stopping: finite fisher information is all you need.
\newblock \emph{arXiv preprint arXiv:2308.12240}, 2023.

\bibitem[Cui et~al.(2023)Cui, Krzakala, Vanden-Eijnden, and Zdeborov{\'a}]{cui2023analysis}
Hugo Cui, Florent Krzakala, Eric Vanden-Eijnden, and Lenka Zdeborov{\'a}.
\newblock Analysis of learning a flow-based generative model from limited sample complexity.
\newblock \emph{arXiv preprint arXiv:2310.03575}, 2023.

\bibitem[De~Bortoli(2022)]{de2022convergence}
Valentin De~Bortoli.
\newblock Convergence of denoising diffusion models under the manifold hypothesis.
\newblock \emph{arXiv preprint arXiv:2208.05314}, 2022.

\bibitem[De~Bortoli et~al.(2021)De~Bortoli, Thornton, Heng, and Doucet]{de2021diffusion}
Valentin De~Bortoli, James Thornton, Jeremy Heng, and Arnaud Doucet.
\newblock Diffusion schr{\"o}dinger bridge with applications to score-based generative modeling.
\newblock \emph{Advances in Neural Information Processing Systems}, 34:\penalty0 17695--17709, 2021.

\bibitem[Derrida(1981)]{derrida1981random}
Bernard Derrida.
\newblock Random-energy model: An exactly solvable model of disordered systems.
\newblock \emph{Physical Review B}, 24\penalty0 (5):\penalty0 2613, 1981.

\bibitem[Donoho et~al.(2000)]{donoho2000high}
David~L Donoho et~al.
\newblock High-dimensional data analysis: The curses and blessings of dimensionality.
\newblock \emph{AMS math challenges lecture}, 1\penalty0 (2000):\penalty0 32, 2000.

\bibitem[Ghio et~al.(2023)Ghio, Dandi, Krzakala, and Zdeborov{\'a}]{ghio2023sampling}
Davide Ghio, Yatin Dandi, Florent Krzakala, and Lenka Zdeborov{\'a}.
\newblock Sampling with flows, diffusion and autoregressive neural networks: A spin-glass perspective.
\newblock \emph{arXiv preprint arXiv:2308.14085}, 2023.

\bibitem[Guth et~al.(2022)Guth, Coste, Bortoli, and Mallat]{guth2022wavelet}
Florentin Guth, Simon Coste, Valentin~De Bortoli, and Stephane Mallat.
\newblock Wavelet score-based generative modeling, 2022.

\bibitem[Haussmann and Pardoux(1986)]{haussmann1986time}
Ulrich~G Haussmann and Etienne Pardoux.
\newblock Time reversal of diffusions.
\newblock \emph{The Annals of Probability}, pages 1188--1205, 1986.

\bibitem[He et~al.(2016)He, Zhang, Ren, and Sun]{Kaiming2015_ResNet}
Kaiming He, Xiangyu Zhang, Shaoqing Ren, and Jian Sun.
\newblock Deep residual learning for image recognition.
\newblock In \emph{2016 IEEE Conference on Computer Vision and Pattern Recognition (CVPR)}, pages 770--778, 2016.
\newblock \doi{10.1109/CVPR.2016.90}.

\bibitem[Ho et~al.(2020)Ho, Jain, and Abbeel]{Ho2020}
Jonathan Ho, Ajay Jain, and Pieter Abbeel.
\newblock Denoising diffusion probabilistic models.
\newblock \emph{Advances in Neural Information Processing Systems}, 2020.

\bibitem[Hyv{\"a}rinen and Dayan(2005)]{hyvarinen2005estimation}
Aapo Hyv{\"a}rinen and Peter Dayan.
\newblock Estimation of non-normalized statistical models by score matching.
\newblock \emph{Journal of Machine Learning Research}, 6\penalty0 (4), 2005.

\bibitem[Kadkhodaie et~al.(2023)Kadkhodaie, Guth, Simoncelli, and Mallat]{kadkhodaie2023generalization}
Zahra Kadkhodaie, Florentin Guth, Eero~P Simoncelli, and St{\'e}phane Mallat.
\newblock Generalization in diffusion models arises from geometry-adaptive harmonic representation.
\newblock \emph{arXiv preprint arXiv:2310.02557}, 2023.

\bibitem[Krizhevsky et~al.(2009)Krizhevsky, Hinton, et~al.]{Krizhevsky2009_CIFAR}
Alex Krizhevsky, Geoffrey Hinton, et~al.
\newblock Learning multiple layers of features from tiny images.
\newblock \emph{Toronto, ON, Canada}, 2009.

\bibitem[Lecun et~al.(1998)Lecun, Bottou, Bengio, and Haffner]{Lecun1998}
Y.~Lecun, L.~Bottou, Y.~Bengio, and P.~Haffner.
\newblock Gradient-based learning applied to document recognition.
\newblock \emph{Proceedings of the IEEE}, 86\penalty0 (11):\penalty0 2278--2324, 1998.
\newblock \doi{10.1109/5.726791}.

\bibitem[Lee et~al.(2022)Lee, Lu, and Tan]{lee2022convergence}
Holden Lee, Jianfeng Lu, and Yixin Tan.
\newblock Convergence for score-based generative modeling with polynomial complexity.
\newblock \emph{arXiv preprint arXiv:2206.06227}, 2022.

\bibitem[Lucibello and M{\'e}zard(2024)]{lucibello2023exponential}
Carlo Lucibello and Marc M{\'e}zard.
\newblock The exponential capacity of dense associative memories.
\newblock \emph{Phys.Rev.Lett}, 132:\penalty0 077301, 2024.

\bibitem[Luo(2022)]{luo2022understanding}
Calvin Luo.
\newblock Understanding diffusion models: A unified perspective.
\newblock \emph{arXiv preprint arXiv:2208.11970}, 2022.

\bibitem[Mezard and Montanari(2009)]{mezard2009information}
Marc Mezard and Andrea Montanari.
\newblock \emph{Information, physics, and computation}.
\newblock Oxford University Press, 2009.

\bibitem[M{\'e}zard et~al.(1987)M{\'e}zard, Parisi, and Virasoro]{mezard1987spin}
Marc M{\'e}zard, Giorgio Parisi, and Miguel~Angel Virasoro.
\newblock \emph{Spin glass theory and beyond: An Introduction to the Replica Method and Its Applications}, volume~9.
\newblock World Scientific Publishing Company, 1987.

\bibitem[Opper and Saad(2001)]{opper2001advanced}
Manfred Opper and David Saad.
\newblock \emph{Advanced mean field methods: Theory and practice}.
\newblock MIT press, 2001.

\bibitem[Paszke et~al.(2019)Paszke, Gross, Massa, Lerer, Bradbury, Chanan, Killeen, Lin, Gimelshein, Antiga, Desmaison, Kopf, Yang, DeVito, Raison, Tejani, Chilamkurthy, Steiner, Fang, Bai, and Chintala]{Pytorch_2019}
Adam Paszke, Sam Gross, Francisco Massa, Adam Lerer, James Bradbury, Gregory Chanan, Trevor Killeen, Zeming Lin, Natalia Gimelshein, Luca Antiga, Alban Desmaison, Andreas Kopf, Edward Yang, Zachary DeVito, Martin Raison, Alykhan Tejani, Sasank Chilamkurthy, Benoit Steiner, Lu~Fang, Junjie Bai, and Soumith Chintala.
\newblock Pytorch: An imperative style, high-performance deep learning library.
\newblock In \emph{Advances in Neural Information Processing Systems 32}, pages 8024--8035. Curran Associates, Inc., 2019.

\bibitem[Poole et~al.(2022)Poole, Jain, Barron, and Mildenhall]{poole2022dreamfusion}
Ben Poole, Ajay Jain, Jonathan~T Barron, and Ben Mildenhall.
\newblock Dreamfusion: Text-to-3d using 2d diffusion.
\newblock \emph{arXiv preprint arXiv:2209.14988}, 2022.

\bibitem[Privman(1990)]{privman1990finite}
Vladimir Privman.
\newblock \emph{Finite size scaling and numerical simulation of statistical systems}.
\newblock World Scientific, 1990.

\bibitem[Raya and Ambrogioni(2023)]{ambrogioni}
Gabriel Raya and Luca Ambrogioni.
\newblock Spontaneous symmetry breaking in generative diffusion models.
\newblock \emph{arXiv preprint arXiv:2305.19693}, 2023.

\bibitem[Ronneberger et~al.(2015)Ronneberger, Fischer, and Brox]{Ronneberger2015_unet}
Olaf Ronneberger, Philipp Fischer, and Thomas Brox.
\newblock U-net: Convolutional networks for biomedical image segmentation.
\newblock In \emph{Medical Image Computing and Computer-Assisted Intervention--MICCAI 2015: 18th International Conference, Munich, Germany, October 5-9, 2015, Proceedings, Part III 18}, pages 234--241. Springer, 2015.

\bibitem[Ruelle(1987)]{ruelle1987mathematical}
David Ruelle.
\newblock A mathematical reformulation of derrida's rem and grem.
\newblock \emph{Communications in Mathematical Physics}, 108:\penalty0 225--239, 1987.

\bibitem[Saharia et~al.(2022)Saharia, Chan, Saxena, Li, Whang, Denton, Ghasemipour, Gontijo~Lopes, Karagol~Ayan, Salimans, et~al.]{saharia2022photorealistic}
Chitwan Saharia, William Chan, Saurabh Saxena, Lala Li, Jay Whang, Emily~L Denton, Kamyar Ghasemipour, Raphael Gontijo~Lopes, Burcu Karagol~Ayan, Tim Salimans, et~al.
\newblock Photorealistic text-to-image diffusion models with deep language understanding.
\newblock \emph{Advances in Neural Information Processing Systems}, 35:\penalty0 36479--36494, 2022.

\bibitem[Sohl-Dickstein et~al.(2015)Sohl-Dickstein, Weiss, Maheswaranathan, and Ganguli]{Sohl_Dickstein2015}
Jascha Sohl-Dickstein, Eric Weiss, Niru Maheswaranathan, and Surya Ganguli.
\newblock Deep unsupervised learning using nonequilibrium thermodynamics.
\newblock In \emph{International Conference on Machine Learning}, 2015.

\bibitem[Song et~al.(2020)Song, Meng, and Ermon]{song2020denoising}
Jiaming Song, Chenlin Meng, and Stefano Ermon.
\newblock Denoising diffusion implicit models.
\newblock \emph{arXiv preprint arXiv:2010.02502}, 2020.

\bibitem[Song and Ermon(2019)]{Song2019}
Yang Song and Stefano Ermon.
\newblock Generative modeling by estimating gradients of the data distribution.
\newblock \emph{Advances in Neural Information Processing Systems}, 2019.

\bibitem[Song et~al.(2021)Song, Sohl-Dickstein, Kingma, Kumar, Ermon, and Poole]{Song_Sohl-Dickstein2021}
Yang Song, Jascha Sohl-Dickstein, Diederik~P Kingma, Abhishek Kumar, Stefano Ermon, and Ben Poole.
\newblock Score-based generative modeling through stochastic differential equations.
\newblock In \emph{International Conference on Learning Representations}, 2021.

\bibitem[Vincent(2011)]{vincent2011connection}
Pascal Vincent.
\newblock A connection between score matching and denoising autoencoders.
\newblock \emph{Neural computation}, 23\penalty0 (7):\penalty0 1661--1674, 2011.

\bibitem[Yang et~al.(2022)Yang, Zhang, Song, Hong, Xu, Zhao, Shao, Zhang, Cui, and Yang]{yang2022diffusion}
Ling Yang, Zhilong Zhang, Yang Song, Shenda Hong, Runsheng Xu, Yue Zhao, Yingxia Shao, Wentao Zhang, Bin Cui, and Ming-Hsuan Yang.
\newblock Diffusion models: A comprehensive survey of methods and applications.
\newblock \emph{arXiv preprint arXiv:2209.00796}, 2022.

\bibitem[Yoon et~al.(2023)Yoon, Choi, Kwon, and Ryu]{yoon2023diffusion}
TaeHo Yoon, Joo~Young Choi, Sehyun Kwon, and Ernest~K Ryu.
\newblock Diffusion probabilistic models generalize when they fail to memorize.
\newblock In \emph{ICML 2023 Workshop on Structured Probabilistic Inference $\{$$\backslash$\&$\}$ Generative Modeling}, 2023.

\bibitem[Yu et~al.(2016)Yu, Seff, Zhang, Song, Funkhouser, and Xiao]{Yu2015_LSUN}
Fisher Yu, Ari Seff, Yinda Zhang, Shuran Song, Thomas Funkhouser, and Jianxiong Xiao.
\newblock Lsun: Construction of a large-scale image dataset using deep learning with humans in the loop, 2016.

\bibitem[Zinn-Justin(2021)]{zinn2021quantum}
Jean Zinn-Justin.
\newblock \emph{Quantum field theory and critical phenomena}, volume 171.
\newblock Oxford university press, 2021.

\end{thebibliography}

\newpage
\appendix
\onecolumn
\begin{center}
    {\LARGE Supplementary Information (SI) Appendix} \\ \vspace{1ex} {\Large Dynamical Regimes of Diffusion Models}
    \\ \vspace{3ex} {\large Giulio Biroli, Tony Bonnaire, Valentin de Bortoli, and Marc M\'ezard}
\end{center}

In this Appendix we first present more details on the analytical frameworks we used to study the backward dynamics of DMs, and then provide additional information on the numerical experiments. As already stressed in the main text, we need specific methods to study the limit of large dimension and large number of data. Such methods have been developed in statistical physics to study dynamics and thermodynamics of a large number of degrees of freedom. We will refer to original articles and, when needed, provide a short-introduction to the methods.

\section{Landau-type expansion for estimating the speciation time}
The speciation transition is similar to the symmetry breaking phenomenon \cite{ambrogioni,GBM} taking place at thermodynamic phase transitions, for instance when a ferromagnetic system develops a non-zero magnetization at low temperature. One way to study this phenomenon is to construct perturbatively the free energy as a function of the field \cite{chaikin1995principles}. In our case, we proceed in a similar way 
by obtaining $\log P_t(\vec x)$ in a perturbative expansion in $e^{-t}$ valid at large times. This approach is justified since speciation takes place at large times. 

We rewrite the probability to be at $\vx$ at time $t$ as
\begin{align}
P_t(\vx)&=\int d\va \; P_0(\va) \frac{1}{\sqrt{2\pi
  \Delta_t}^d}\exp\left(-\frac{1}{2} \frac{(\vx-\va
  e^{-t})^2}{\Delta_t}\right)\nonumber\\
  &= \frac{1}{\sqrt{2\pi
  \Delta_t}^d} \exp\left(-\frac{1}{2} \frac{\vx^2}{\Delta_t}+g(\vx) \right) 
\end{align}
where $g(\vx)$, defined as 
\begin{align}
g(\vx)=\log \int d\va \; P_0(\va) \exp\left(-\frac{1}{2} \frac{\va
  ^2 e^{-2t}}{\Delta_t}\right)\exp\left(\frac{e^{-t}\vx.\va}{\Delta_t}\right)
\end{align}
can be interpreted in a field-theoretic (or probabilistic) approach, as a generative function for connected correlations of the variables $\va$ \cite{zinn2021quantum}. Expanding this function at large times, one finds 
\begin{align}
g(\vx)= \frac{e^{-t}}{\Delta_t} \sum_{i=1}^d x_i\langle a_i\rangle+\frac{1}{2}\frac{e^{-2t}}{\Delta_t^2}\sum_{i,j=1}^d
x_ix_j[\langle a_i a_j\rangle-\langle a_i\rangle \langle a_j\rangle] +O\left((x e^{-t})^3\right)
\end{align}
where the brackets $\langle.\rangle$ denote the expectation value with respect to the effective distribution 
$ P_0(\va) e^{-\va^2 e^{-2t}/(2\Delta_t) } $. Using this expansion, one finds  at large times:
\begin{align}
\log P_t(\vx)= C+ \frac{e^{-t}}{\Delta_t} \sum_{i=1}^d x_i\langle a_i\rangle - 
\frac{1}{2\Delta_t} \sum_{i,j=1}^d
x_i M_{ij} x_j+O\left((x e^{-t})^3\right)
\end{align}
where $C$ is an $\vx$-independent term and
\begin{align}
M_{ij}= \delta_{ij}- e^{-2t} [\langle a_i a_j\rangle-\langle a_i\rangle \langle a_j\rangle] 
\end{align}
The curvature of $\log P_t(\vx) $ depends on the spectrum of the matrix $M$. At large times $M$ is close to the identity, all its eigenvalues are positive. The  shape changes qualitatively at the largest time $t_S$ such that the largest eigenvalue of $M$ crosses zero. The speciation time is characterized by a change of curvature of the effective potential  $-\log P_t(\vx)$.

Notice that the effective measure used for computing the correlations of $a_i$ variables appearing in the definition of $M$
can be substituted at large times by the original measure $P_0(\va)$. Therefore $M\simeq \mathbb{I}-e^{-2t} \hat C_0$ where $\hat C_0$ is the covariance of the distribution $P_0$, which can be estimated as the covariance of the data. This leads to the (second) general criterion discussed in the main text for the speciation transition.

\section{Gaussian mixtures: speciation time and asymptotic stochastic processes in the high-dimensional limit }
We consider the case when $P_0(\vec a)$ is the superposition of two Gaussian clusters of equal weight, with means $\pm \vec m$, and the same variance $\sigma^2$.
We ensure that the two Gaussians are well separated in the limit of large dimensions, by assuming that  $|\vec m|^2 =d\tilde \mu^2$, with  $\sigma$ and $\tilde \mu$  of order 1. In the following we present the detail of the analysis of the backward dynamics in the limit of large dimension. In the study of regimes I and II, we shall assume that $P_t^e(\vx)$ coincides with its population counterpart $P_t(\vx)$, and use the latter. We justify this assumption in the analysis of regime III, by showing that for $\alpha=\frac{\log n}{d}$ finite, the speciation time, and hence regime I and II, take place before the collapse, i.e. when $P_t^e(\vx)$ coincides with its population counterpart $P_t(\vx)$.  

\subsection{Spectral estimate of $t_S$}
The speciation time computed from the eigenvalue criterion of the previous section is easily computed in this case. The matrix $M$ is given by
$M_{ij}=(1-\sigma^2 e^{-2t}) \delta_{ij}-e^{-2t} m_i m_j$
and its largest eigenvalue is $(1-\sigma^2 e^{-2t}- d\tilde \mu^2 e^{-2t})$. We get therefore in the large $d$ limit $t_S=\frac{1}{2}\log(d\tilde\mu^2)$
which up to subleading corrections identifies the speciation timescale as $$t_S=\frac{1}{2}\log(d).$$

\subsection{Asymptotic stochastic process in regime I and symmetry breaking}
In the large dimensional limit, we can provide a full analytic study of the dynamics in regime I, i.e. at the beginning of the backward process following \cite{GBM}). 

As shown in Sect.~\ref{sect_collapseGM} of this Appendix, when one studies the dynamics at times $t>t_C$  the empirical distribution at time $t$, $P_t^e(\vx)$ is well approximated by $P_t(\vx)$ which is the convolution of the initial distribution $P_0$ (a mixture of Gaussians centered at $\pm \vec m$) and the diffusion kernel proportional to $e^{-(\vx-\va e^{-t})^2/2}$. 
The explicit expression is thus 
\begin{align}
    P_t(\vx) = \frac{1}{2\sqrt{2\pi\Gamma_t}^d}\left[
e^{-(\vx-\vm e^{-t})^2/(2 \Gamma_t)}
+e^{-(\vx+\vm e^{-t})^2/(2 \Gamma_t)}
\right]
\end{align}
where $\Gamma_t=\sigma^2 e^{-2 t}+ \Delta_t$ goes to $1$ at large times.
The $\log$ of this probability is
\begin{align}
   \log P_t(\vx) = -\frac{\vx^2}{2\Gamma_t}+\log \cosh\left( \vec x \cdot \vec m \;\frac{e^{-t}}{\Gamma_t}\right)
\end{align}
and hence the score reads 
\begin{align}\label{eq:scoregm}
   S_i(\vx) = -\frac{x_i}{\Gamma_t}+m_i \frac{e^{-t}}{\Gamma_t}\tanh\left( \vec x \cdot \vec m \; \frac{e^{-t}}{\Gamma_t}\right)
\end{align}

At the beginning of the backward process (for $t\gg t_S$), the $x_i$ are i.i.d. Gaussian variables with unit variance. In consequence the overlap with the centers of the Gaussian model, $\vec m\cdot \vec x(t)$, scales as $\sqrt{d}$ in the large dimensional limit. Strictly speaking, regime I is defined as the scaling regime (or the time-window) in which 
$\vec m\cdot \vec x(t)$ keeps this scaling with $d$. Introducing 
the quantity $q(t)=\frac{1}{\sqrt{d}} \vec m\cdot \vec x(t)$ and using the notation $t_S=(1/2)\log d$, one can therefore write the backward equation on each $x_i$ as:
\begin{align}\label{eq:regimeigm}
-dx_i=x_i+2\left(-\frac{x_i}{\Gamma_t}+m_i \frac{e^{-t}}{\Gamma_t}\tanh\left(q(t) \frac{e^{-(t-t_S)}}{\Gamma_t}\right)\right)dt +d\eta_i(t) \,
\end{align}
where $d\eta_i(t)$ is square root of two times the Brownian motion. 
This shows that in regime I, on the timescale $t_S$, each $x_i$ satisfies a Langevin equation in which the interactions with all the other variables is through the fluctuating rescaled overlap $q(t)$. It is a kind of dynamical mean-field equation \cite{bonnaire2023high}. 

In order to obtain the equation on $q(t)$, one can sum the equation on all $x_i$s.
One therefore finds that this projection of the trajectory point $\vx(t)$
on the direction linking  the centers of the two Gaussians in the mixture satisfies the closed backward stochastic differential equation cited in the main text:
\begin{align}
    -dq=-\frac{\partial V(q,t)}{\partial q}dt+d\eta(t)
\end{align}
where $d\eta(t) $ is square root of two times the Brownian motion, and the potential $V(q,t)$ reads 
\begin{align}
    V(q,t)=\frac{1}{2}q^2-2\tilde \mu^2 \log \cosh\left(q  e^{-t} \sqrt{d}\right)
\end{align}
Clearly this potential is quadratic at times $t\gg (1/2) \log d$, and it develops a double well structure when $t\ll (1/2) \log d$. Equation (\ref{eq:qt}) is a Langevin equation in an inverted potential $V(q,t)$. As explained in the main text, the resulting dynamics leads to the symmetry breaking in which trajectories commit to one of the two classes.  
The scaling variable controlling such phenomenon is $e^{-\sqrt{t}}d$, as also found by the other means explained in the main text and this Appendix. 

Note that the time where the curvature at $q=0$ vanishes,
$ \frac{\partial^2 V }{\partial q^2}(q=0,t^*)=0$,  is 
\begin{align}
    t^*=\frac{1}{2}\log(2d\tilde\mu^2).
\end{align}
In the main text we identify the speciation time with $t_S=\frac{1}{2}\log (d \tilde \mu^2)$, or equivalently $e^{-2t_S}d \tilde \mu^2=1$ but any other choice of constant different from one, e.g. $1/2 $ as above, would be equivalent in the large $d$ limit  (it gives subleading correction to $t_S$ in the large $d$ limit).

\subsection{Analytic computation of the speciation time defined from cloning}
We now consider the criterion for speciation used in numerical experiments, namely the probability $\phi(t)$ that two trajectories, cloned at time $t$, end at time $0$ in the same class. For gaussian mixtures,  this can be computed as 
follows.
\begin{align}
P(t)=\left[\int_{\vec x_1 \cdot \vm >0,\vec x_2 \cdot \vm >0} +\int_{\vec x_1 \cdot \vm <0,\vec x_2 \cdot \vm<0}\right] d\vec x_1 d\vec x_2 d\vec y \,\left(P(\vec x_1, 0|\vec y,t)P(\vec x_2, 0|\vec y,t)P(\vec y,t)\right)
\end{align}
where $P(\vec x_1, 0|\vec y,t)$ is the probability that the backward process ends in $\vec x_1$ knowing that it was in $\vec y$ at time $t$. This conditional probability can be rewritten as $P(\vec x_1, 0|\vec y,t)=P(\vec x_1, 0,\vec y,t)/P(\vec y,t)$. We can now use the forward process to compute all the probabilities involved, using 
$P(\vec x_1, 0,\vec y,t)=P(\vec y,t|\vec x_1, 0)P(\vec x_1, 0)$:
\[
P(\vec x_1, 0,\vec y,t)=P_0(\vec x_1)\frac{e^{-\frac{(\vec y -\vec x_1 e^{-t})^2}{2\Delta_t}}}{\sqrt{2\pi\Delta_t}^d}
\]
hence leading to:
\begin{align}
P(\vec x_1, 0|\vec y,t)=P_0(\vec x_1)\frac{e^{-\frac{(\vec y -\vec x_1 e^{-t})^2}{2\Delta_t}}}{\sqrt{2\pi\Delta_t}^d}\frac{1}{P(\vec y,t)}
\end{align}
 
We now insert this expression into the equation for $P(t)$ and integrate over $\vec x_1$ and $\vec x_2$. The integrals can be done by noticing that this is equivalent to drawing both $x_1$ and $x_2$ from one Gaussian with probability $1/4$ and from the other Gaussian with probability $1/4$. One thus finds:
\begin{align}
\phi(t)=\int \frac{d\vec y}{2\sqrt{2\pi(\sigma^2 e^{-2t}+\Delta_t)}^d}\; 
 \frac{
e^{-2\frac{(\vec y -\vec \mu e^{-t})^2}{2(\sigma^2 e^{-2t}+\Delta_t)}}
+
e^{-2\frac{(\vec y +\vec \mu e^{-t})^2}{2(\sigma^2 e^{-2t}+\Delta_t)}}
}{
e^{-\frac{(\vec y -\vec \mu e^{-t})^2}{2(\sigma^2 e^{-2t}+\Delta_t)}}+
e^{-\frac{(\vec y +\vec \mu e^{-t})^2}{2(\sigma^2 e^{-2t}+\Delta_t)}}
}
\end{align}
All the integrals on the components orthogonal to $\vm$ factorize and can be done: they give one. One ends up with a one-dimensional integral equal to:
\begin{align}
\phi(t)=\frac{1}{2} \int dy \frac{G(y,m e^{-t},\Gamma_t)^2+G(y,-m e^{-t},\Gamma_t)^2 }{G(y,m e^{-t},\Gamma_t)+G(y,-m e^{-t},\Gamma_t)}\ ,
\end{align}
where $G(y,u,v)$ is a Gaussian probability density function for the real variable $y$, with mean  $u$ and variance $v$, $m=|\vec m|=\tilde \mu \sqrt{d}$, $\Gamma_t= \Delta_t+\sigma^2 e^{-2t}$. This  one-dimensional integral is easily done numerically. For large $d$, the probability $\phi(t)$
that the two clones end up in the same cluster is a decreasing function of $t$, going from $\phi(0)=1$ to $\phi(\infty)=1/2$, and the speciation time for this Gaussian mixture model, estimated from $\phi (t_s)=.775$ is of order $t_s\simeq (1/2)\log d$ in the limit of large dimensions.

\subsection{Asymptotic stochastic process in regime II}
At the end of regime I the overlap of $\vec{x}(t)$ with the centers of the Gaussian model diverge: for some trajectories $q(t)$ goes to plus infinity, for others to minus infinity. This corresponds to a change of scaling regime. In fact, beyond regime I, the overlap $\vec{x}(t)\cdot \vec{m}$ scales proportionnally to $d$. 

After the speciation transition, trajectories are committed to a given center. Conditioning on trajectories which correspond to the center $+\vec m$, and using that $\vec{x}(t)\cdot \vec{m} \rightarrow +\infty$, the score (\ref{eq:scoregm}) simplifies to:  
\begin{align}
   S_i^+(\vx) = -\frac{x_i}{\Gamma_t}+m_i \frac{e^{-t}}{\Gamma_t}
\end{align}
This is the score of the backward process of a single Gaussian centered in $+\vec m$. The resulting equation on $x_i$ is therefore: 
\begin{align}\label{eq:regimeiigm}
-dx_i=\left(-\frac{x_i}{\Gamma_t}+m_i \frac{e^{-t}}{\Gamma_t}\right)dt +d\eta_i(t)
\end{align}
where $d\eta_i(t)$ is square root of two times a Brownian motion. This equation is the one of the backward process of the single Gaussian centered in $+\vec m$, and therefore guarantees that all trajectories evolving with such equation will generate the single Gaussian centered in $+\vec m$. 

We have focused on the trajectories committed to the center $+\vec m$. One can proceed analogously for the ones committed to the center $-\vec m$. The results above still holds with $m_i$ replaced by $-m_i$.
\\
We conclude this section on regime I and II by stressing that we expect that our analysis can be generalized to other models. The Curie-Weiss already studied in \cite{GBM} is an example. The dynamics in regime II has also been studied in \cite{ghio2023sampling} using a model for the score.

\section{Gaussian Mixtures: collapse time} \label{sect_collapseGM}

\subsection{Setting and methods} 
In order to study the structure of $ P_t^e(\vx)$ around a point $\vx=\va^1e^{-t}+\vec z\sqrt{\Delta_t}$, where $\vec z$ has i.i.d. Gaussian distributed components with mean zero and variance one,  we start from the representation $P_t^e(\vx)= \left[Z_1+Z_{2...n}\right]/\sqrt{2\pi
  \Delta_t}^d$
  where $Z_1= e^{-(\vx-\va^1 e^{-t})^2/(2\Delta_t)}= e^{-(\vec z^2)/2}  $
and
\begin{align}
Z_{2...n}= \sum_{\mu=2}^n e^{-(\vx-\va^\mu e^{-t})^2/(2\Delta_t)} 
\end{align}

We now study $P_t(\vx)$ in the limit of large $d$, large $n$, keeping $\alpha=\log n/d$ fixed. The first piece behaves as $Z_1\simeq e^{-d/2}$. In the second piece, $Z_{2...n}$,
we assume, without loss of generality, that the first $n_+$ data points were sampled from the Gaussian  with mean $\vec m$, and the last $n_-=n-n_+$ ones were sampled from the Gaussian  with mean $-\vec m$. In the large $n$ limit, because of the law of large numbers $n_+/n$ goes to 1/2.
We can write $Z_{2...n}=Z_++Z_- $,where
\begin{align}
Z_+&= \sum_{\mu=2}^{n_+} e^{-(\vx-\va^\mu e^{-t})^2/(2\Delta_t)} \\
Z_-&= \sum_{\mu=n_++1}^{n} e^{-(\vx-\va^\mu e^{-t})^2/(2\Delta_t)}
\end{align}
$Z_1$, $Z_+$ and $Z_-$ give the contribution to $P_t(\vx)$ from, respectively, the data point $\mu=1$ (which is supposed to have been generated from the gaussian with mean $\vec m$), the other $n_+-1$ data points generated from this same gaussian with mean $\vec m$, and the $n_-$ data points generated from the gaussian with mean $-\vec m$. The two quantities $Z_\pm$ are partition functions and we shall show that, for typical values of 
$\vx=\va^1e^{-t}+\vec z\sqrt{\Delta_t}$, the free energies $(1/d)\log Z_\pm$ concentrates in the large $d$ limit around a value $\psi_\pm(t)$. Furthermore, $\psi_-<\psi_+$. Therefore in the large $d$ limit the collapse transition is identified as the time $t_c$ such that $\psi_+(t_c)=-1/2$.

As already pointed out in the main text, in order to obtain $Z_\pm$, standard concentration methods, e.g. central limit theorem, do not apply as each term of the sum corresponds to the exponential of a random variable scaling as $\log n$ \cite{ben2005limit}. $Z_\pm$ correspond to the partition function of a system with $n-1$ independent `random energy levels'.  This is some elaboration of the `Random Energy Model' which was introduced originally in \cite{derrida1981random} as a simple model of glass transition, and studied in the limit we are interested in, i.e. $n,d$ going to infinity with $\alpha=\frac{\log n}{d}$ fixed. This model has played a central role in the physics of glassy systems \cite{mezard1987spin}. It was first solved with methods of theoretical physics \cite{derrida1981random}. It was then thoroughly studied by probabilistic rigorous methods \cite{ruelle1987mathematical}. 

For a mathematical introduction that make connections with physics and computer science see the book \cite{mezard2009information}. For our purposes, it is also useful the recent work \cite{lucibello2023exponential} where a similar problem was addressed in the context of dense associative memories.

\subsection{Computation of $\psi_+$}
Given a time $t$, $Z_+$ is the partition function of a system with $n_+-1$ independent 'energy levels' $E^\mu =(\vx-\va^\mu e^{-t})^2/(2\Delta_t) $ at thermal equilibrium at a temperature $1$.  All these energies are independent and identically distributed, and they are of order $d$. Writing $\ve^\mu=E^\mu/d$, we denote by $\rho_t(\ve)$ the probability density function from which  these reduced energies are drawn independently. This density $\rho_t(\ve) $ satisfies a large deviation principle with a rate function $f_t(\ve)$:
\begin{align}
    \rho_t(\ve)=e^{-d f_t(\ve)}
\end{align}
We define 
 $g_t(\lambda)$ the Legendre transform $g_t(\lambda)=\max_{\ve} [-f_t(\ve)-\lambda \ve]$ and use 
\begin{align}
e^{d g_t(\lambda)}= \int d\ve \; e^{-d [f_t(\ve)+\lambda\ve] }=\expect_+ e^{-\lambda (\vx- \va e^{-t})^2/2\Delta_t}
\label{gtdef}
\end{align}
where $\expect_+ $ is an expectation on $\vec a$ drawn from the gaussian measure of mean $\vm$ and variance $\sigma^2$. 
The right-hand side of (\ref{gtdef}) is the convolution of two gaussians which is easily computed. Using the fact that $(\vx-\vm e^{-t})^2/d$ concentrates at large $d$ around $\sigma_t^2+\Delta_t$, where $\sigma_t^2=\sigma^2 e^{-2 t}$, we get:
\begin{align}
g_t(\lambda)=\frac{1}{2}\log \frac{\Delta_t}{\Delta_t+\lambda \sigma_t^2}- \frac{1}{2}\frac{\lambda (\Delta_t+\sigma_t^2)}{\Delta_t+\lambda \sigma_t^2}
\end{align}
The Legendre transform can be done analytically. the maximum of $g_t(\lambda)+\lambda
 \ve$ is obtained at 
\begin{align}
    \lambda^*(\ve)= \frac{\sigma_t^2-4\Delta_t\ve +\sqrt{\sigma_t^4+8\Delta_t^2 \ve +8\Delta_t\sigma_t^2 \ve}}{4\sigma_t^2 \ve}
\end{align}
The inverse transform is obtained from the maximum of $-f_t(\ve)-\lambda \ve$ which is obtained at 
\begin{align}
\ve^*(\lambda)= -\frac{d g_t}{d\lambda }= \frac{\Delta_t^2+2\Delta_t\sigma_t^2+\lambda \sigma_t^2}{2(\Delta_t+\lambda \sigma_t^2)^2}
\end{align}
and the rate function is

\begin{align}
 f_t(\ve)=
\frac{\Delta_t}{2\sigma_t^2 (\sigma_t^2+A)}\left[
-8\Delta_t\ve +(1-6 \ve)\sigma_t^2+(1+2 \ve A)
\right]
-\frac{1}{2}\log\frac{4\Delta_t\ve}{\sigma_t^2+A}
\label{fdef}
\end{align}
where
\begin{align}
    A=\sqrt{\sigma_t^4+8\ve \Delta_t (\Delta_t+\sigma_t^2)}
\end{align}
The partition function is 
$Z_+= \sum_{\mu=2}^{n_+} e^{-d \ve^\mu}$.
The annealed approximation to this partition function is 
$Z_+^{ann}=n_+\int d \ve \; e^{-d (\ve+f_t(\ve)) }$. Using Laplace's method, we see that this integral is dominated by $\ve=\ve^*(\lambda=1)=1/2$, and gives 
\begin{align}
\psi_+^{ann}=\lim_{d\to\infty} \frac{1}{d}  \log Z_+^{ann} =
\alpha-\frac{1}{2}-f_t\left(\frac{1}{2}\right)
=\alpha+g_t(1)= \alpha+\frac{1}{2}\log\frac{\Delta_t}{\Delta_t+\sigma_t^2}-\frac{1}{2}
\end{align}
The methods for analyzing random energy models are standard (see \cite{derrida1981random, mezard2009information}). For a presentation close to the developments which we use here, see \cite{lucibello2023exponential},  in particular its Appendix A. Using either with first and second moment methods, or with replicas, we find that the annealed free energy is exact, namely $\psi_+=\psi_+^{ann}$ when $t<t_{cond}$, but at $t>t_{cond}$ a condensation phenomenon occurs, the partition function is dominated by the levels with lowest energy, $\ve_{min}$ given by the smallest root of 
$\alpha=\hat f_t(\ve)$. The condensation time $t_{cond}$ is  thus characterized by the fact that $\ve_{min}=1/2$, which gives:
\begin{align}
\alpha=f_{t_{cond}}(1/2)
\end{align}
So the final expression for the free energy of this random energy model is:
\begin{align}
\psi_+=\lim_{d\to\infty} \frac{1}{d}\log Z_+ &= 
\alpha+\frac{1}{2}\log\frac{\Delta_t}{\Delta_t+\sigma_t^2}-\frac{1}{2}
\ \ \  \text{if}\ \  t<t_{cond}
\nonumber \\
&= -\frac{1}{2} \ \ \  \text{if}\ \  t>t_{cond}
\label{phiplus}
\end{align}
and its condensation transition is
\begin{align}
t_{cond}=\frac{1}{2}\log\left[1+\frac{\sigma^2}{n^{2/d}-1}
\right]
\end{align}

\subsection{Computation of $\psi_-$}
Given $t$, the partition function $Z_-$ is again a partition function of a random energy model, which can be studied following exactly the same steps as for $Z_+$. The computation of $g_t(\lambda)$ is slightly different. Now, $(\vx-\vm e^{-t})^2/d$ concentrates at large $d$ around $4(\vm^2/d)e^{-2t}+\sigma_t^2+\Delta_t$+ This gives: 
\begin{align}
g_t(\lambda)=\frac{1}{2}\log \frac{\Delta_t}{\Delta_t+\lambda \sigma_t^2}- \frac{1}{2}\frac{\lambda [M_t +\Delta_t+\sigma_t^2]}{\Delta_t+\lambda \sigma_t^2}
\end{align}
where $M_t= 4 (\vm^2/d) e^{-2t}$
We now list the changes in the subsequent formulations, keeping the same notations as in the previous section. 
The maximum of $g_t(\lambda)+\lambda
 \ve$ is obtained at 
\begin{align}
    \lambda^*(\ve)= \frac{\sigma_t^2-4\Delta_t\ve +\sqrt{\sigma_t^4+8\Delta_t^2 \ve +8\Delta_t\sigma_t^2 \ve+8 \Delta_t M_t\ve}}{4\sigma_t^2 \ve}
\end{align}
The inverse transform is obtained from the maximum of $-f_t(\ve)-\lambda \ve$ which is at
\begin{align}
\ve^*(\lambda)= -\frac{d g_t}{d\lambda }= \frac{\Delta_t^2+2\Delta_t\sigma_t^2+\lambda \sigma_t^2+M_t\Delta_t
}{2(\Delta_t+\lambda \sigma_t^2)^2}
\end{align}
and the rate function is

\begin{align}
 f_t(\ve)=
\frac{\Delta_t\left[
-8\Delta_t\ve +(1-6 \ve)\sigma_t^2+(1+2 \ve A)
\right]
+M_t\left[ 
\sigma_t^2-8\ve \Delta_t+A
\right]
}{2\sigma_t^2 (\sigma_t^2+A)}
-\frac{1}{2}\log\frac{4\Delta_t\ve}{\sigma_t^2+A}
\end{align}
where
\begin{align}
    A=\sqrt{\sigma_t^4+8\ve \Delta_t (\Delta_t+\sigma_t^2)+8 \Delta_t M_t\ve}
\end{align}
The annealed approximation to the partition function is 
$Z_-^{ann}=n_-\int d \ve \; e^{-d (\ve+f_t(\ve)) }$. This integral is dominated by \begin{align}
\ve=\ve^*(\lambda=1)=\frac{1}{2} \left[1+\frac{\Delta_t M_t}{(\Delta_t+\sigma_t^2)^2}\right]
\end{align}
and gives 
\begin{align}
\psi_-^{ann}=\lim_{d\to\infty} \frac{1}{d}  \log Z_-^{ann} 
=\alpha+g_t(1)= \alpha+\frac{1}{2}\log\frac{\Delta_t}{\Delta_t+\sigma_t^2}-\frac{1}{2}- \frac{1}{2}\frac{M_t}{\Delta_t+\sigma_t^2}
\end{align}

The annealed free energy is exact when $t<t_{cond}$. The condensation takes place at $t>t_{cond}$, then the partition function is dominated by the levels with lowest energy, $\ve_{min}$ given by the smallest root of 
$\alpha=\hat f_t(\ve)$. The condensation time $t_{cond}$ is  thus characterized by the fact that $\ve_{min}=\ve^*(\lambda=1)$, which gives:
\begin{align}
\alpha=f_{t_{cond}}\left( \ve=
\frac{1}{2} \left[1+\frac{\Delta_t M_t}{(\Delta_t+\sigma_t^2)^2}\right]
\right)
\end{align}
So the final expression for the free energy of this random energy model is:
\begin{align}
\psi_-=\lim_{d\to\infty} \frac{1}{d}\log Z_- &= 
\alpha+\frac{1}{2}\log\frac{\Delta_t}{\Delta_t+\sigma_t^2}-\frac{1}{2}- \frac{1}{2}\frac{M_t}{\Delta_t+\sigma_t^2}
\ \ \  \text{if}\ \  t<t_{cond}\nonumber \\
&= -\frac{1}{2} \left[1+\frac{\Delta_t M_t}{(\Delta_t+\sigma_t^2)^2}\right] \ \ \  \text{if}\ \  t>t_{cond}
\end{align}
and its condensation transition is
\begin{align}
t_{cond}=\frac{1}{2}\log\left[1+\frac{\sigma^2}{n^{2/d}-1}
\right]
\end{align}
It is easy to see that at any time $\psi_-<\psi_+$.

\subsection{Final expression}

Going back to the representation $P_t^e(\vx)= \left[Z_1+Z_{2...n}\right]/\sqrt{2\pi
  \Delta_t}^d$,
  where $Z_1= e^{-(\vx-\va^1 e^{-t})^2/(2\Delta_t)}= e^{-(\vec z^2)/2}  $, it is made of three contributions:
\begin{enumerate}
\item
$Z_1$ which behaves at large $d$ as $e^{-d/2}$
\item 
 $Z_+ $ which behaves at large $d$ as $e^{-d \psi_+}$
\item 
 $Z_- $ which behaves at large $d$ as $e^{-d \psi_-}$
\end{enumerate}

In the limit of large dimensions $d\to\infty$, the contribution from $Z_-$ is irrelevant because $\psi_-<\psi_+$. The comparison of the two other terms depends on whether $\psi_+$ is larger or smaller than $-1/2$. We thus find two time regimes separated by a collapse transition time $t_C$.
\begin{itemize}
    \item at large times, $t >t_C$, $\psi_+>-(1/2)$ and the probability $P_t^e(\vx)$ is dominated by the term $Z_+$. This is the regime which is not collapsed.
    \item at small  times, $t <t_C$, $\psi_+<-(1/2)$ and the probability $P_t^e(\vx)$ is dominated by the term $Z_1= e^{-(\vx-\va^1 e^{-t})^2/(2\Delta_t)}$. When used in the backward diffusion, this gives a score that attracts $\vx$ towards $\va^1$ at short times.
\end{itemize}

We have thus shown that, in the backward process,  trajectories which are at time $t$ at typical points of the form 
$\vx=\va^1 e^{-t}+\vec z\sqrt{\Delta_t}$ are  attracted towards $\va^1$ if $t<t_C$. This identifies $t_C$ as the collapse time.

Looking at the explicit expression 
(\ref{phiplus}) 
of $\psi_+$, we see that the collapse time $t_C$ is identical to the condensation time of the random energy model, $t_{cond}$. This result, derived exactly in this case of gaussian mixtures, is actually a general result, as discussed in the main paper. It could also be shown using the REM approach, provided the distribution of energies satisfies a large deviation principle.

\section{Details on the numerical experiments}

\subsection{Gaussian mixtures} \label{sect:exp_GM}

\paragraph*{Speciation experiment.} In the case of a mixture of two Gaussian clusters centered on $\pm \vec m \in \mathbb{R}^d$ with variance $\sigma^2$, the score function $\vec \cF(\vx,t)$ can be analytically expressed as
\begin{equation}
    \vec \cF(\vx(t), t) = \vec m \frac{e^{-t}}{\Gamma_t} \tanh\left(\frac{e^{-t}}{\Gamma_t} \vx(t) \cdot \vec m \right) - \frac{\vx(t)}{\Gamma_t},
\end{equation}
where $\Gamma_t = \Delta_t + \sigma^2 e^{-2t}$, with $\Delta_t = 1 - e^{-2t}$.
We can then discretize the stochastic differential equation associated to the backward process as
\begin{equation}
    \vx(t+1) = \vx(t) + \eta \left[ \vx(t) + 2 \cF(\vx(t), t)\right] + \vec \xi\sqrt{2\eta},
\end{equation}
where $\vec \xi \sim \mathcal{N}(0, I)$, $\vec x_T \sim \mathcal{N}(0, I)$ and $\eta = T/L$, with $T=10$ the time horizon and $L = 1000$ the number of discrete steps.
Two clones $\vec x_1(t)$ and $\vec x_2(t)$ share the same trajectory until a given time below which they have independant noise realizations $\vec \xi$. At the end of the backward diffusion, one can then recover the class of the clone $i$ by projecting it onto $\vec m$ as $\mathrm{sign}\left(\vec m \cdot \vec x_{i}(0)\right)$. In the experiment corresponding to Fig.~2 of the main text, we use $\vec m = \left[1, \ldots, 1\right]$, $\sigma^2 = 1$, and each point is obtained by averaging over $10\,000$ initial conditions. \\

\paragraph*{Collapse experiment.} The main component of the collapse experiment is the excess entropy density $f(t)$ which vanishes for $t \leq t_C$. When dealing with Gaussian mixtures, $f(t)$ can be derived exactly and we additionally establish a strategy to approximate it numerically as follows. At a given time $t$, we draw $n'$ samples $\vec x^{(\nu)}(t) = e^{-t} a^{(\mu)} + \sqrt{1-e^{-2t}} \vec \xi^{(\nu)}$, where $\mu$ is chosen uniformly in $\{1, \ldots, n\}$. The entropy $s(t) = -\int \mathrm{d} \vec x \; P_t(\vec x) \log P_t(\vec x)$ can then be approximated as the empirical average over the $n'$ samples, leading to the empirical estimate $f^e(t)$ of the excess entropy, using
\begin{align}
s^e(t)=-\frac{1}{n'd}\sum_{\nu=1}^{n'}\log P^e_t\left(\vec x_t^{(\nu)}\right),
\end{align}
where $P_t^e(\vec x)$ is given as usual in terms of the original dataset by
\begin{align}
P_t^e(\vec x)=\frac{1}{n}\sum_{\mu=1}^n \frac{1}{\sqrt{2\pi \Delta_t}^d}\exp\left(-\frac{(\vec x-\vec a^\mu e^{-t})^2}{2\Delta_t}\right).
\end{align}
This approximation is valid for $t>t_C$, where $P_t(\vec x) \approx P_t^e(\vec x)$.
As shown in Fig.~2 of the main text where we fix $n=20\,000$ and $n'=250\,000$ for several $d$, this estimate fits quite remarkably the analytical curve. Such a procedure therefore allows for the numerical derivation of the collapse $t_C$ from the dataset only, under the assumption that one uses the  exact empirical score.

\subsection{Realistic datasets} \label{sect:exp_real}

\paragraph*{Denoising Diffusion Probabilistic Models.}
In the second part of the paper, we learn the score by training a Denoising Diffusion Probabilistic Model (DDPM) in discrete time, as introduced by \cite{Ho2020}. In this context, the forward process has a variance schedule $\{\beta_t'\}_{t'=1}^L$, where $L$ is the time horizon given as a number of steps, fixed to 1000 in our experiments. In our case, the variance is evolving linearly from $\beta_1 = 10^{-4}$ to $\beta_{1000} = 2 \times 10^{-2}$.
A sample at timestep $t'$, denoted $\vec x(t')$ can therefore be expressed readily from its initial state, $\vec x(0) = \vec a$, as
\begin{equation}
    \vec x(t') = \sqrt{\overline{\alpha}(t')} \vec a + \sqrt{1 - \overline{\alpha}(t')} \vec \xi(t'),
\end{equation}
where $\overline{\alpha}(t') = \prod_{s=1}^{t'} (1-\beta_s)$ and $\vec \xi$ is a standard and centered Gaussian noise. This equation in fact corresponds to the discretization of the Ornstein-Uhlenbeck Eq. (1) given in the main text under the following reparameterization of the timestep $t'$,
\begin{equation}
    t = -\frac{1}{2} \log \left( \overline{\alpha}(t')\right),
\end{equation}
where $t$ is the time as defined in the main text.
We then train a neural network to learn $\vec \xi_\theta\left(\vec x(t'), t'\right)$, the noise at time $t'$ by iteratively optimizing the loss function
\begin{equation}
    \mathcal{L}(\theta) = \lVert \vec \xi(t') - \vec \xi_\theta\left(\vec x(t'), t'\right) \rVert^2.
\end{equation}
Once learned, the denoiser can then be used to generate a new sample as
\begin{equation}
    \vec x(t'-1) = \frac{1}{\sqrt{1-\beta_{t'}}} \vec x(t') - \frac{\beta_{t'}}{\sqrt{\left(1 - \beta_{t'}\right) \left(1-\overline{\alpha}_{t'}\right)}} \vec \xi_{\theta}\left(\vec x(t'), t'\right) + \sqrt{\beta_{t'}} \vec z,
\end{equation}
where $t'$ runs from $T$ to one, $\vec x(T) \sim \mathcal{N}(0, I)$, and $\vec z \sim \mathcal{N}(0, I)\delta(t' > 1)$, which, in the continuous limit, is equivalent to the score-based diffusion studied in the main text \cite{luo2022understanding}, where $-\vec \xi_{\theta}\left(\vec x(t'), t'\right)/\sqrt{1 - \overline{\alpha}_{t'}}$ corresponds to the score.
\\

\paragraph*{Datasets and preprocessing.}
Our experiments are based on five datasets: MNIST \cite{Lecun1998}, CIFAR \cite{Krizhevsky2009_CIFAR}, ImageNet16/32 \cite{Chrabaszcz2017_Imagenet_downsampled}, and LSUN \cite{Yu2015_LSUN}. For each dataset, we focus only on two well-disctinct and balanced classes that are: zero and seven for MNIST ($n=10\,000$), horses and cars for CIFAR ($n=3000$), lesser pandas and seashores for ImageNet16 and 32 ($n=2000$), and churches and conference rooms for LSUN ($n=40\,000$ or $n=200$). In the case of MNIST, the images are zero-padded from their original size to $32\times 32$. For LSUN, images are center-cropped at size $256\times 256$ before being resized to $64\times 64$. All the datasets but MNIST have three color channels that are kept for the training, and the data are centered in each channel. The different datasets therefore allow to span several values of $d$ ranging from $1024$ (for MNIST) to $12\,288$ for LSUN. The values of $\Lambda$ given in the main text corresponding to the largest eigenvalue of the covariance matrix are computed from the first channel only. \\

\begin{figure*}[!htb]
\centering
\includegraphics[width=1\linewidth]{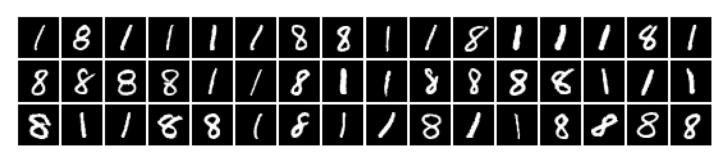}
\caption{MNIST one and eight generated samples with $n=10\,000$.}
\label{fig:samples_MNIST}  

\centering
\includegraphics[width=1\linewidth]{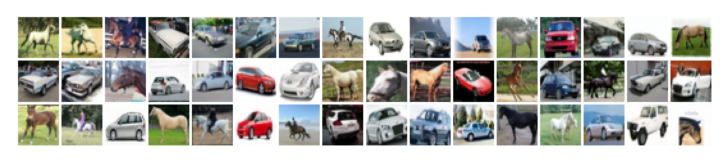}
\caption{CIFAR horses and cars generated samples with $n=3000$.}
\label{fig:samples_CIFAR}  

\centering
\includegraphics[width=1\linewidth]{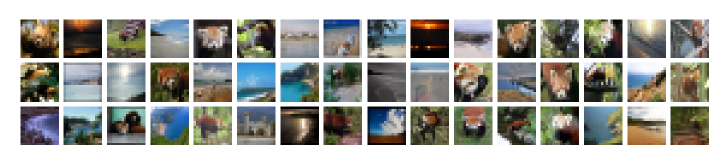}
\caption{ImageNet16 lesser pandas and seashores generated samples with $n=2000$.}
\label{fig:samples_I16}  

\centering
\includegraphics[width=1\linewidth]{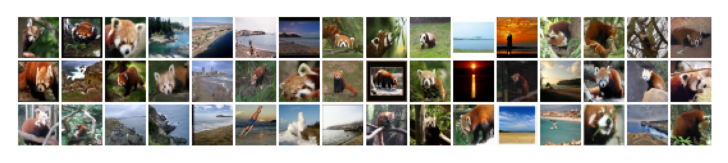}
\caption{ImageNet32 lesser pandas and seashores generated samples with $n=2000$.}
\label{fig:samples_I32}  
\end{figure*}

\begin{figure}[!htb]
\begin{subfigure}{.5\textwidth}
    \centering
    \includegraphics[width=\linewidth]{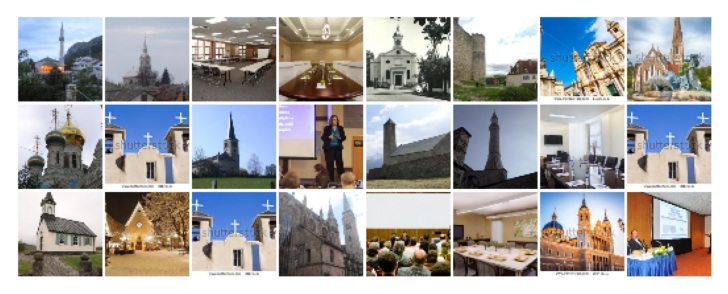}
    \caption{}
\end{subfigure} \quad
\begin{subfigure}{.5\textwidth}
    \centering
    \includegraphics[width=\linewidth]{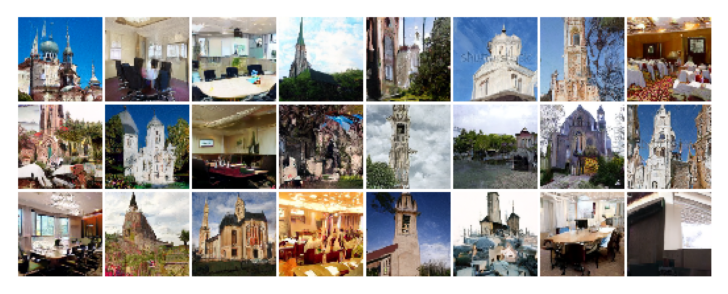}
    \caption{}
\end{subfigure}
\caption{LSUN churches and conference rooms generated samples with (a) $n= 200$ or (b) $n=40\,000$.}
\label{fig:samples_LSUN}
\end{figure}

\paragraph*{Architecture and training.} For all the datasets, the denoiser $\vec \xi_\theta\left(\vec x(t'), t'\right)$ has a U-Net architecture similar to previous applications of DDPMs \cite{Ho2020, song2020denoising}. More precisely, it implements four resolution levels, each having two convolutional residual blocks with group normalization, and self attention is applied to the two intermediary levels. The time is added into each block through a standard sinusoidal position embedding, resulting in a total of $25.7$ million parameters. The models are then trained using ADAM optimizer with a fixed learning rate of $10^{-4}$ for $350$k steps with batches of size $128$, except for LSUN for which we reduced it to $64$. All the $n$ images were used to train the models, without any data augmentation.
It is worth mentioning that the results exposed in the main text do not depend on the formal architecture of the denoiser, nor the optimization scheme, as long as the network is able to learn a score close to the true one (or equivalently, a sufficiently good denoiser). In Figs.~\ref{fig:samples_MNIST} to~\ref{fig:samples_LSUN}, we show several samples generated from our six trained models. \\

\paragraph*{Speciation experiment.} To estimate numerically the speciation time from the probability $\phi(t)$ that the two clones end up in the same class, we need to classify the generated samples. To do so, we train the PyTorch \cite{Pytorch_2019} implementation of the ResNet-18 architecture \cite{Kaiming2015_ResNet} with pre-trained weights on ImageNet. One model is trained for each of the dataset using all the $n$ samples. The resulting classifiers yield $95$\% test accuracy at worst on LSUN, and more than $99$\% for the other datasets. \\

\begin{figure}[!htb]
\begin{subfigure}{.5\textwidth}
    \centering
    \includegraphics[width=.9\linewidth]{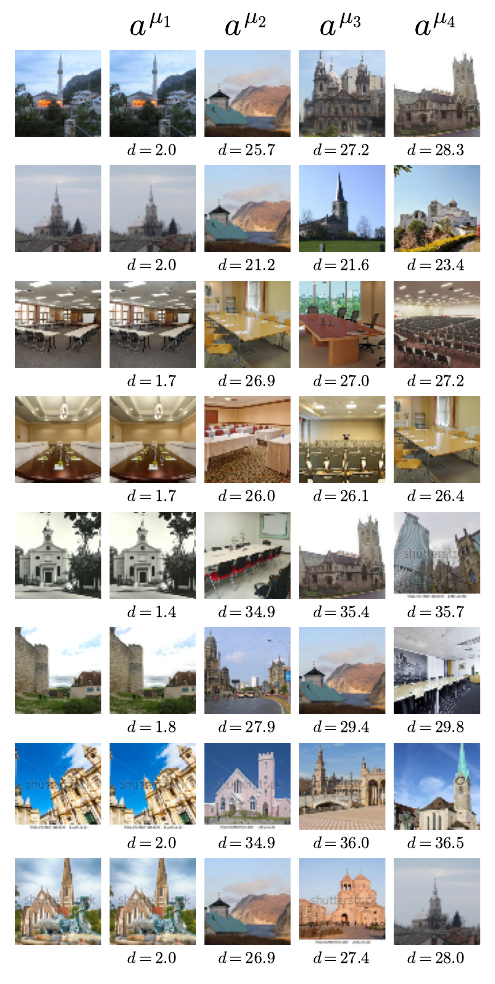}
    \caption{}
    \label{fig:Collapse_LSUN_200}
\end{subfigure} \quad
\begin{subfigure}{.5\textwidth}
    \centering
    \includegraphics[width=.9\linewidth]{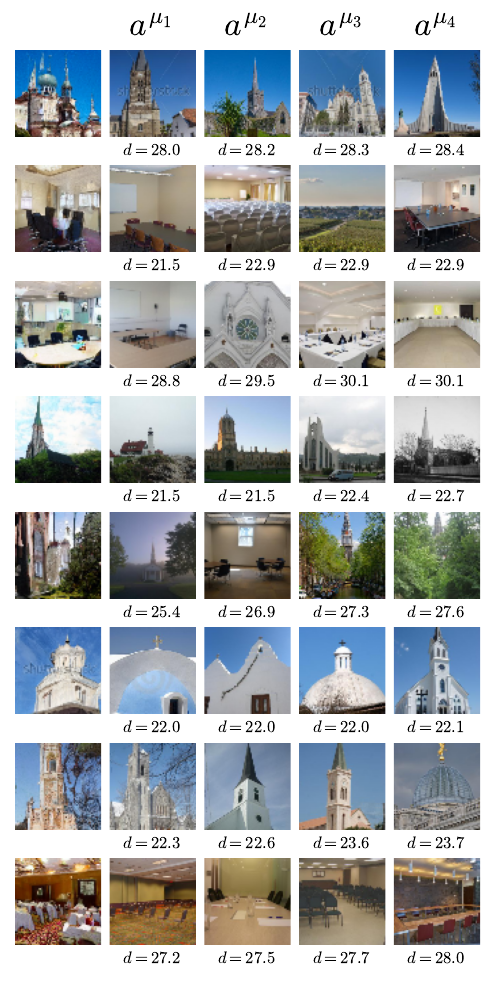}
    \caption{}
    \label{fig:Collapse_LSUN_40k}
\end{subfigure}
\caption{Illustration of the collapse on the LSUN dataset trained with (a) $n=200$ or (b) $n = 40\,000$ datapoints. Each row corresponds to a generated sample shown in the left-most column. The other columns display the first four nearest neighbors in the training set, denoted as $a^{\mu_1}, a^{\mu_2}, a^{\mu_3}$, and $a^{\mu_4}$. Below each of them is indicated the $L_2$ norm $d$ to the generated sample.}
\label{fig:Collapse_LSUN}
\end{figure}

\paragraph*{Collapse experiment.} To study the collapse, we devise two methods. First, we exploit the cloned trajectories but we now focus on the indices of the nearest neighbors in the training set at the end of the backward process. Using a small number of training data ($n = 2000$ for ImageNet and $n = 200$ for LSUN) ensures the backward dynamics to collapse onto one of the training datapoint. When $t < t_C$, the two clones have the same nearest neighbor, with a very small Euclidean distance, showing that the collapse has indeed taken place. This approach allows us to compute $\phi_C(t)$, the probability that the two clones collapse on the same datapoint. We illustrate this phenomenon on the LSUN dataset trained with either $n=200$ or $n=40\,000$ in Fig.~\ref{fig:Collapse_LSUN}.
For a random batch of generated samples, we show the four nearest neighbors $\{a^{\mu_1}, a^{\mu_2}, a^{\mu_3}, a^{\mu_4}\}$ (ordered by increasing distance) in the training set. In the case of a model trained with a small number of datapoints, the generated images collapse onto $\mu_1$, as shown in Fig.~\ref{fig:Collapse_LSUN_200}. The model being fixed, one solution to avoid the collapse is to increase $n$, resulting in the generation of new samples, as illustrated in Fig.~\ref{fig:Collapse_LSUN_40k}.
In a second approach, we exploit the indices of nearest neighbors in the training set \emph{during} the backward process. At any times $t < t_C$, the index of the nearest neighbor $\mu_1(t)$ is therefore fixed. At $t=0$, the distance $d_{\mu_1}$ to the nearest neighbor is vanishing, as seen before. The average of this distribution for $n=4000$ generated samples yields the estimate $\hat{t}_C$.
The entropy appearing in $f^e(t)$ is computed the same way as in Gaussian mixtures (see Sect.~\ref{sect:exp_GM}), making use of the training sets and the forward process, with $n'=400\,000$.

\end{document}